%% file: acm_main.tex
\documentclass[format=acmsmall,screen]{acmart}
\pdfoutput=1 

\makeatletter
\if@ACM@authorversion
\usepackage{fancyhdr}
\AtBeginDocument{%
    \addtolength{\footskip}{2.0\baselineskip}%
    \fancyfoot[L]{\textit{Pre-print}}%
}
\fi
\makeatother

\usepackage[utf8]{inputenc}
\usepackage[T1]{fontenc}
\usepackage[USenglish]{babel}
\usepackage{microtype}

\input{metadata/hypersetup}

\usepackage{subcaption}
\usepackage{xcolor,xurl}
\usepackage{tikz,relsize}
\usetikzlibrary{shapes.geometric, shapes.multipart, arrows, arrows.meta, positioning, calc, chains, backgrounds}
\usepackage{mathtools,xfrac,dsfont,bm}
\usepackage{cleveref}
\usepackage{float, adjustbox}
\usepackage{booktabs}
\usepackage{enumitem}

\setlist{itemsep=0.15em,parsep=0em,topsep=0.5em}
\newcommand{\examplefigurewidth}{\linewidth}
\input{figures/examples_macro}

\DeclareMathOperator*{\argmax}{argmax}
\DeclareMathOperator*{\argmin}{argmin}

\newcommand{\ifacm}[1]{}
\newcommand{\ifarxiv}[1]{#1}
\newcommand{\measure}[1]{\hyperref[sec:measures-of-interpretability]{\emph{#1}}}
\newcommand{\motivation}[1]{\hyperref[sec:motivation-for-interpretability]{\emph{#1}}}
\newcommand{\category}[1]{\hyperref[sec:introduction:abstraction-level]{\emph{#1}}}
\newcommand{\intrinsic}[1]{\hyperref[sec:introduction:intrinsic-v-post-hoc]{\emph{#1}}}
\newcommand{\posthoc}[1]{\hyperref[sec:introduction:intrinsic-v-post-hoc]{\emph{#1}}}
\newcommand{\type}[2]{\hyperref[#1]{\emph{#2}}}
\newcommand{\method}[2]{\hyperref[#1]{\emph{#2}}}

\setcopyright{rightsretained}
\acmJournal{CSUR}
\acmYear{2022} \acmVolume{55} \acmNumber{5} \acmArticle{155} \acmMonth{12} \acmDOI{10.1145/3546577}
\copyrightyear{2022}

\begin{document}

\title{Post-hoc Interpretability for Neural NLP: A Survey}

\author{Andreas Madsen}
\email{andreas.madsen@mila.quebec}
\orcid{0000-0002-1487-2796}
\additionalaffiliation{%
    \institution{École Polytechnique de Montréal}
    \streetaddress{2500 Chem. de Polytechnique}
    \city{Montréal}
    \state{Quebec}
    \country{Canada}
}
\author{Siva Reddy}
\email{siva.reddy@mila.quebec}
\orcid{0000-0003-3753-0323}
\additionalaffiliation{%
    \institution{McGill University}
    \streetaddress{845 Sherbrooke St W}
    \city{Montréal}
    \state{Quebec}
    \country{Canada}
}
\additionalaffiliation{%
    \institution{Facebook CIFAR AI Chair}
}
\author{Sarath Chandar}
\email{sarath.chandar@mila.quebec}
\orcid{0000-0002-9678-2830}
\authornotemark[1]
\additionalaffiliation{%
    \institution{Canada CIFAR AI Chair}
}
\affiliation{%
    \institution{Mila}
    \streetaddress{6666 Rue Saint-Urbain}
    \city{Montréal}
    \state{Quebec}
    \country{Canada}
}

\begin{abstract}
\input{chapters/abstract}
\end{abstract}


\maketitle

\begin{acks}
\input{chapters/acknowledgement}
\end{acks}

\input{chapters/introduction}
\input{chapters/motivation}
\input{chapters/motivating_example}
\input{chapters/measures}

\input{chapters/methods}
\input{chapters/input_features}
\input{chapters/adversarial_examples}
\input{chapters/influential_examples}
\input{chapters/counterfactuals}
\input{chapters/natural_language}
\input{chapters/concepts}
\input{chapters/vocabulary}
\input{chapters/ensemble}
\input{chapters/linguistic_information}
\input{chapters/rules}

\input{chapters/limitations}

\input{chapters/findings}
\input{chapters/future_directions}
\input{chapters/conclusion}

\bibliographystyle{acm/ACM-Reference-Format}
\bibliography{bibliography}

\clearpage
\ifacm{
\appendix
\section{Input features}
\label{sec:appendix:start}

\input{chapters/input_features_ig}
\input{chapters/input_features_shap}
\input{chapters/input_features_anchors}
\section{Adversarial examples}

\input{chapters/adversarial_examples_sea}
\section{Counterfactuals}

\input{chapters/counterfactuals_polyjuice}
\section{Rules}
\label{sec:appendix:end}
\input{chapters/rules_compositional}
}
\end{document}

The main LaTeX file is `acm_main.tex`. Supplementary material is included after `\appendix`.

There are no special requirements.

%% file: metadata/hypersetup.tex
\hypersetup{breaklinks,
            pdfauthor={Andreas Madsen, Siva Reddy, Sarath Chandar},
            pdftitle={Post-hoc Interpretability for Neural NLP: A Survey},
            pdfsubject={A survey on post-hoc interpretability for neural networks applied to natural language processing (NLP). The survey discusses interpretability methods, such as LIME, Integrated Gradient, Influence Functions, PCA or t-SNE projection, Linguistic Probing, etc. Each interpretability method is categorized by how they communicate. How each interpretability method is evaluated in terms of functional-groundedness / faithfulness and human-groundedness is regularly discussed. Finally, the survey describes future directions for interpretability research in NLP.},
            pdfkeywords={Survey, Neural Networks, Natural Language Processing, NLP, Interpretability, Post-hoc, Overview, Comparison}}

%% file: chapters/abstract.tex
Neural networks for NLP are becoming increasingly complex and widespread, and there is a growing concern if these models are responsible to use.
Explaining models helps to address the safety and ethical concerns and is essential for accountability.
Interpretability serves to provide these explanations in terms that are understandable to humans. Additionally, post-hoc methods provide explanations after a model is learned and are generally model-agnostic.
This survey provides a categorization of how recent post-hoc interpretability methods communicate explanations to humans, it discusses each method in-depth, and how they are validated, as the latter is often a common concern.

%% file: chapters/acknowledgement.tex
SC and SR are supported by the Canada CIFAR AI Chairs program and the NSERC Discovery Grant.

%% file: chapters/introduction.tex
\section{Introduction}

Large neural NLP models, most notably BERT-like models \citep{Devlin2019,Liu2019,Brown2020}, have become highly widespread, both in research and industry applications \citep{Wolf2019}.
This increase of model complexity is motivated by a general correlation between model size and test performance \citep{Kaplan2020, Brown2020}.
Due to their immense complexity, these models are generally considered black-box models.
A growing concern is therefore if it is responsible to deploy these models.

Concerns such as safety, ethics, and accountability are particularly important when machine learning is used for high-stakes decisions, such as healthcare, criminal justice, finance, etc. \citep{Rudin2019}, including NLP-focused applications such as translation, dialog systems, resume screening, search, etc. \citep{Doshi-Velez2017}.
For many of these applications, neural models have been shown to exhibit unwanted biases and similar ethical issues \citep{Rudin2019, Obermeyer2019, Brown2020, Bender2021, Mehrabi2021, Garrido-Munoz2021}.

\citet{Doshi-Velez2017a} argue, among others \citep{Lipton2018}, that these ethical and safety issues stem from an ``incompleteness in the problem formalization''. 
While these issues can be partially prevented with robustness and fairness metrics, it is often not possible to consider all failure modes. Therefore, quality assessment should also be done through model explanations.
Furthermore, when models do fail in critical applications, explanations must be provided to facilitate the accountability process. Providing these explanations is often a core motivation for interpretability. In \Cref{sec:motivation-for-interpretability} we provide aditional motivating factors.

\citet{Doshi-Velez2017a} define \emph{interpretability} as the ``ability to explain or to present in understandable terms to a human''. However, what constitutes as an ``understandable'' explanation is an interdisciplinary question.
An important work from social science by \citet{Miller2019}, argues that \emph{effective explanations} must be selective in the sense one must select ``one or two causes from a sometimes infinite number of causes''. Such observation necessitates organizing interpretability methods by how and what they selectively communicate.

This survey presents such an organization in \Cref{tab:overview}, where each row represents a communication approach. For example, the first row describes \emph{input feature} explanations that communicate what tokens are most relevant for a prediction. In general, each row is ordered by how abstract the communication approach is, although this is an approximation. Organizing by the method of communication is discussed further in \Cref{sec:introduction:abstraction-level}. 

\begin{table}[H]
\centering
\resizebox{\examplefigurewidth}{!}{\input{figures/overview-table}}
\caption[Overview of \posthoc{post-hoc} interpretability methods]{Overview of \posthoc{post-hoc} interpretability methods, where § indicates the section the method is discussed. Rows describe how the explanation is communicated, while columns describe what information is used to produce the explanation. The order of both rows and columns indicates level of abstraction and amount of information, respectively. However, this order is only approximate.

Furthermore, because this survey focuses on \posthoc{post-hoc} methods, the \intrinsic{intrinsic} section of this table is incomplete and merely meant to provide a few comparative examples. The specifc \intrinsic{intrinsic} methods shown are: \emph{Attention} \citep{Bahdanau2015}, \emph{GEF} \citep{Liu2019a}, \emph{NILE} \citep{Kumar2020}. \emph{Prototype Networks} and \emph{Auxiliary Task} refer to types of models.

\textsuperscript{$\mathcal{C}$}: Depends on checkpoints during training. \textsuperscript{$\mathcal{D}$}: Depends on supplementary dataset. \textsuperscript{$\mathcal{H}$}: Depends on second-order derivative. \textsuperscript{$\mathcal{M}$}: Depends on supplementary model. \textsuperscript{${\dagger}$}: Depends only on dataset and white-box access.}
\label{tab:overview}
\end{table}

Each interpretability method uses different kinds of information to produce its explanation, in \Cref{tab:overview} this is indicated by the columns\footnote{\emph{Black-box}: the method only evaluates the model. \emph{Dataset}: the method has access to all training and validation observations. \emph{Gradient}: the gradient of the model is computed. \emph{Embeddings}: the method uses the word embedding matrix. \emph{White-box}:  the method knows everything about the model, such as all weights and all operations. However, the method is not specific to a particular architecture. \emph{Model specific}: the method is specific to the architecture. Note, neural model in NLP are usually differentiability and have an embedding matrix. We therefore do not consider these properties constraints.}. The columns are ordered by an increasing level of information. Again, this is an inexact ranking but serves as a useful tool to contrast the methods.

\Cref{tab:overview} frames the overall structure of this survey. Where each method section from \ref{sec:input-features} to \ref{sec:rules} covers a row of \Cref{tab:overview}. However, first we cover motivation (\cref{sec:motivation-for-interpretability}), how to validate interpretability  (\cref{sec:measures-of-interpretability}), and a motivating example (\cref{sec:motivating-example}). The method sections can be read somewhat independently but will refer back to these general topics. 

In contrast to other surveys and tutorials on interpretability methods \citep{Wallace2020a,Molnar2019,Belinkov2020,Chakraborty2017,Bhatt2020,Sun2021,Danilevsky2020,Belinkov2019,Tjoa2020} which only discusses the most popular approaches (usually 3 among \type{sec:input-features}{input features}, \type{sec:adversarial-examples}{adversarial examples}, \type{sec:influential-examples}{influential examples}, \method{sec:vocabulary:projection}{projection}, and \type{sec:linguistic-information}{linguistic information}), this survey offers a more diverse overview of communication approaches. We hope this leads to more questioning about how we communicate. Additionally, we consistently comment on how each method is validated (\measure{groundedness}), an important discussion we find is often missing.

Finally, the survey limits itself to \posthoc{post-hoc} interpretability methods. These are methods that provide their explanation after a model is trained and are often model-agnostic. This is in contrast to \intrinsic{intrinsic} methods, where the model architecture itself helps to provide the explanation. These terms  are described further in \Cref{sec:introduction:intrinsic-v-post-hoc}.

\subsection{Organizing by method of communication}
\label{sec:introduction:abstraction-level}

As a categorization of communication strategies, it's standard in the interpretability literature to distinguish between methods that explain a single observation, called \category{local explanations}, and methods that explain the entire model called \category{global explanations} \citep{Doshi-Velez2017a, Adadi2018, Carvalho2019, Molnar2019, Chatzimparmpas2020, Bhatt2020}. In this survey, we also consider an additional category of methods that explains an entire output-class, which we call \category{class explanations}.

To subdivide these categories further, \Cref{tab:overview} orders each communication strategy by their abstraction level. As an example, see \Cref{fig:introduction:explanation-examples}, where an \type{sec:input-features}{input features} explanation highlights the input tokens that are most responsible for a prediction; because this must refer to specific tokens, its ability to provide abstract explanations is limited. For a highly abstract explanation, consider the \type{sec:natural-language}{natural language} category which explains a prediction using a sentence and can therefore use abstract concepts in its explanation.

\begin{figure}[h]
    \centering
    \examplefigure{introduction}
    \caption{Fictive visualization of an \type{sec:input-features}{input features} explanation which highlights tokens and a \type{sec:natural-language}{natural language} explanation, applied on a sentiment classification task \citep{Wang2019}. $y = \mathtt{pos}$ means the gold label is \textit{positive} sentiment.}
    \label{fig:introduction:explanation-examples}
\end{figure}

Communication methods that have a higher abstraction level are typically easier to understand (more \measure{human-grounded}), but the trade-off is that they may reflect the model's behavior less (less \measure{functionally-grounded}). Because the purpose of interpretability is to communicate the model to a human, this trade-off is necessary \citep{Rudin2019, Miller2019}. Which communication strategy should be used must be decided by considering the applications and to whom the explanation is communicated to. In \Cref{sec:measures-of-interpretability} we discuss \measure{human-groundedness} and \measure{functionally-groundedness} in-depth and how to measure them, such that an informed decision can be made.

\Cref{tab:overview} does have some limitations. Firstly, ordering explanation methods by their abstraction level is an approximation, and while \category{global explanations} are generally more abstract than \category{local explanations} this is not always true. For example, the explanation ``simply print all weights'' (not included in \Cref{tab:overview}), is arguably the lowest possible abstraction level, however it's also a \category{global explanation}. Secondly, there are explanation categories that are not included, such as \emph{intermediate representations}. This category of explanation depends on models that are \intrinsic{intrinsically} interpretable, which are not the subject of this paper.
We elaborate on this in \Cref{sec:introduction:intrinsic-v-post-hoc}.

\subsection{Intrinsic versus post-hoc interpretability}
\label{sec:introduction:intrinsic-v-post-hoc}

A fundamental motivation for interpretability is accountability. For example, if a predictive mistake happens which caused harm, it is important to explain why this mistake happened \citep{Doshi-Velez2017}. Similarly, for high-stakes decisions, it is important to minimize the risk of model failure by explaining the model before deployment \citep{Rudin2019}. In other words, it is important to distinguish between when interpretability is applied proactively or retroactively to the model's deployment.

It is standard in the literature to categorize if an interpretability method can be applied retroactively or proactively. Unfortunately, the terminology for this taxonomy is not standardized \citep{Carvalho2019}. This survey focuses on the methods that can be applied retroactively, for which the term \posthoc{post-hoc} is used. Similarly, we use the term \intrinsic{intrinsic} to refer to models that are interpretable by design. These terms were chosen as the best compromise between established terminology \citep{Molnar2019, Du2019, Jacovi2020} and correctness in terms of their dictionary definition.



\paragraph{\intrinsic{Intrinsic} methods} These inherently depend on models that by design are interpretable. Because of this relation, it is also often referred to as \emph{white-box models} \citep{Rudin2019, Du2019, Chatzimparmpas2020}. However, the term \emph{white-box} is slightly misleading, as it is often only a part of the transparent model.

As an example, consider \emph{intermediate representation} explanations, this category depends on a model that is constrained to produce a meaningful \emph{intermediate representation}. In Neural Modular Networks \citep{Andreas2016, Gupta2020} this could be \texttt{find-max-num(filter(find()))}, which represents how to extract an answer from a question-paragraph-pair. However, how this representation is produced is not necessarily \intrinsic{intrinsically} interpretable.


\intrinsic{Intrinsic} methods are attractive because they may be more responsible to use in high-stakes decision processes. However, as \citet{Jacovi2020} argue, ``a method being \emph{inherently interpretable} is merely a claim that needs to be verified before it can be trusted''. Verifying this is often non-trivial, as has repeatedly been shown with \emph{Attention} \citep{Bahdanau2015}, where multiple papers have found contradicting conclusions regarding interpretability \citep{Wiegreffe2020,Jain2019,Serrano2019,Vashishth2019}. 

\paragraph{\posthoc{Post-hoc} methods} These are the focus of this paper. While many \posthoc{post-hoc} methods are model-agnostic, this is not a necessary property, and in some cases does only apply to a category of models. Indeed, in this paper, only methods that apply to neural networks are discussed.

Because of the inherent ability to explain the model after training, \posthoc{post-hoc} methods are valuable in legal proceedings, where models may need to be explained retroactively \citep{Doshi-Velez2017}. Additionally, they fit into existing quality assessment structures, such as those used to regulate banking, where quality assessment is also done after a model has been built \citep{Bhatt2020}. Finally, it is guaranteed that they will not affect model performance.

However, \posthoc{post-hoc} methods are often criticized for providing false explanations, and it has been questioned if it is reasonable to expect models, that were not designed to be explained, to be explained anyway \citep{Rudin2019}. This is a valid concern, however producing \intrinsic{intrinsic} methods is often very task dependent and therefore a difficult process which is rarely done in the industry \citep{Bhatt2020}. \posthoc{Post-hoc} method are often much more adaptable and their impact can therefore be much greater if they can provide accuate explanations. The question of how to validate explanations is therefore very important and is covered in detail in \Cref{sec:measures-of-interpretability}. Furthermore, we pay special attention to how each method is validated in the literature throughout the survey.

\paragraph{Comparing} Both \intrinsic{Intrinsic} and \posthoc{post-hoc} methods have their merits, but often provide different values in terms of \motivation{accountability}. Finally, \posthoc{post-hoc} methods can often be applied also to \intrinsic{intrinsicly} interpretable models. Observing a correlation between methods from these two categories can therefore provide validation of both methods \citep{Jain2019}.

%% file: figures/overview-table.tex
\begin{tikzpicture}[
    every text node part/.style={align=center},
    category/.style={rectangle, fill=white, anchor=center, xshift=-0.5mm},
    title/.style={rectangle, fill=black!2, anchor=center},
    content/.style={rectangle, fill=black, opacity=1, fill opacity=0.06, text opacity=1, anchor=center},
    >={Latex[width=1.5mm,length=1.5mm]},
]
\large
\pgfmathsetmacro{\tmargin}{0.1}
\pgfmathsetmacro{\tboxwidth}{2.5}
\pgfmathsetmacro{\tboxheight}{0.985}
\pgfmathsetmacro{\tcategoryheight}{0.3}
\pgfmathsetmacro{\tcategoryrelfontsize}{-1}

\newcommand{\pref}[2]{#1 \hyperref[#2]{§\,\ref*{#2}}}

\newcommand{\rAs}{\rBe + \tmargin} \newcommand{\rAe}{\rAs + \tboxheight}
\newcommand{\rBs}{\rCe + \tmargin} \newcommand{\rBe}{\rBs + \tboxheight}
\newcommand{\rCs}{\rDe + \tmargin} \newcommand{\rCe}{\rCs + \tboxheight}
\newcommand{\rDs}{\rEe + \tmargin} \newcommand{\rDe}{\rDs + \tboxheight}
\newcommand{\rEs}{\rFe + \tmargin + \tcategoryheight + \tmargin} \newcommand{\rEe}{\rEs + \tboxheight}
\newcommand{\rFs}{\rGe + \tmargin + \tcategoryheight + \tmargin} \newcommand{\rFe}{\rFs + \tboxheight}
\newcommand{\rGs}{\rHe + \tmargin} \newcommand{\rGe}{\rGs + \tboxheight}
\newcommand{\rHs}{\rIe + \tmargin} \newcommand{\rHe}{\rHs + \tboxheight}
\newcommand{\rIs}{\rJe + \tmargin} \newcommand{\rIe}{\rIs + \tboxheight}
\newcommand{\rJs}{\tmargin} \newcommand{\rJe}{\rJs + \tboxheight}

\newcommand{\cAs}{\tmargin} \newcommand{\cAe}{\cAs + \tboxwidth}
\newcommand{\cBs}{\cAe + \tmargin} \newcommand{\cBe}{\cBs + \tboxwidth}
\newcommand{\cCs}{\cBe + \tmargin} \newcommand{\cCe}{\cCs + \tboxwidth}
\newcommand{\cDs}{\cCe + \tmargin} \newcommand{\cDe}{\cDs + \tboxwidth}
\newcommand{\cEs}{\cDe + \tmargin} \newcommand{\cEe}{\cEs + \tboxwidth}
\newcommand{\cFs}{\cEe + \tmargin + \tmargin} \newcommand{\cFe}{\cFs + \tboxwidth}

\newcommand{\rtitle}[2]{
    \fill[title] (-\tmargin-\tboxwidth, \csname r#1s\endcsname) rectangle node{\vphantom{$^i_ a$}#2\vphantom{$^i_a$}} (-\tmargin, \csname r#1e\endcsname);
}
\newcommand{\rcategory}[2]{
    \fill[category] (-\tmargin-\tboxwidth, \csname r#1e\endcsname + \tmargin + \tcategoryheight) rectangle node{\relsize{\tcategoryrelfontsize} #2} (-\tmargin, \csname r#1e\endcsname + \tmargin);
    \draw (\cAs, \csname r#1e\endcsname + \tmargin) -- (\cEe, \csname r#1e\endcsname + \tmargin);
    \draw (\cFs, \csname r#1e\endcsname + \tmargin) -- (\cFe, \csname r#1e\endcsname + \tmargin);
}

\newcommand{\ctitle}[2]{
    \fill[title] (\csname c#1s\endcsname, \rAe + \tmargin + \tmargin) rectangle node{\vphantom{$^i_ a$}#2\vphantom{$^i_a$}} (\csname c#1e\endcsname, \rAe + \tmargin + \tmargin + \tboxheight);
}
\newcommand{\ccategory}[2]{
    \fill[category] (\csname c#1s\endcsname, \rAe + \tmargin + \tmargin + \tboxheight + \tmargin + \tcategoryheight) rectangle node{\relsize{\tcategoryrelfontsize} #2} (\csname c#1e\endcsname, \rAe + \tmargin + \tmargin + \tboxheight + \tmargin);
    \draw (\csname c#1s\endcsname - \tmargin, \rJs) -- (\csname c#1s\endcsname - \tmargin, \rAe);
}

\newcommand{\tcontent}[4]{
    \fill[content] (\csname c#2s\endcsname, \csname r#1s\endcsname) rectangle node{\vphantom{$^i_ a$}#4\vphantom{$^i_a$}} (\csname c#3e\endcsname, \csname r#1e\endcsname);
}

\draw[->] (\cAs, \rAe + \tmargin + \tmargin + \tboxheight + \tcategoryheight + \tmargin + \tmargin) -- (\cFe, \rAe + \tmargin + \tmargin + \tboxheight + \tcategoryheight + \tmargin + \tmargin) node[above left, xshift=-0.2cm]{less information \hspace{9.3cm} more information};

\draw[->] (\cAs - \tmargin - \tmargin - \tboxwidth - \tmargin, \rAe) -- (\cAs - \tmargin - \tmargin - \tboxwidth - \tmargin, \rJs) node[anchor=mid, below, rotate=-90, xshift=-6cm]{lower abstraction \hspace{5cm} higher abstraction};

\rcategory{A}{local explanation}
\rtitle{A}{input \\ features}
\rtitle{B}{adversarial \\ examples}
\rtitle{C}{influential \\ examples}
\rtitle{D}{counter-\\factuals}
\rtitle{E}{natural \\ language}
\rcategory{F}{class explanation}
\rtitle{F}{concepts}
\rcategory{G}{global explanation}
\rtitle{G}{vocabulary}
\rtitle{H}{ensemble}
\rtitle{I}{linguistic \\ information}
\rtitle{J}{rules}

\ccategory{A}{post-hoc \hspace{0.9cm}}
\ctitle{A}{black-box}
\ctitle{B}{dataset}
\ctitle{C}{gradient}
\ctitle{D}{embeddings}
\ctitle{E}{white-box}
\ccategory{F}{intrinsic \hspace{1.1cm}}
\ctitle{F}{model specific}

\tcontent{A}{A}{A}{\pref{SHAP}{sec:input-features:shap}}
\tcontent{A}{A}{B}{\hspace{2.1cm}\pref{LIME}{sec:input-features:lime},\\\hspace{2.1cm}\pref{Anchors}{sec:input-features:anchors}}
\tcontent{A}{C}{C}{\pref{Gradient}{sec:input-features:gradient},\\ \pref{IG}{sec:input-features:integrated-gradient}}
\tcontent{A}{F}{F}{Attention}
\tcontent{B}{A}{A}{\pref{SEA$^\mathcal{M}$}{sec:adversarial-examples:sea}}
\tcontent{B}{C}{D}{\pref{HotFlip}{sec:adversarial-examples:hotflip}}
\tcontent{C}{B}{C}{\pref{Influence Functions$^\mathcal{H}$}{sec:influential-examples:influence-functions}\\\pref{TracIn$^\mathcal{C}$}{sec:influencial-examples:tracin}}
\tcontent{C}{B}{E}{\hspace{4.3cm} \pref{Representer Pointers$^{\dagger}$}{sec:influential-examples:representer-point-selection}}
\tcontent{C}{F}{F}{Prototype \\ Networks}
\tcontent{D}{A}{A}{\pref{Polyjuice$^{\mathcal{M},\mathcal{D}}$\\}{sec:counterfactuals:polyjuice}}
\tcontent{D}{A}{C}{\pref{MiCE$^\mathcal{M}$}{sec:counterfactuals:mice}}
\tcontent{E}{A}{A}{\pref{CAGE$^{\mathcal{M},\mathcal{D}}$\\}{sec:natural-language:cage}}
\tcontent{E}{F}{F}{GEF$^{\mathcal{D}}$, NILE$^\mathcal{D}$}

\tcontent{F}{E}{E}{\pref{NIE$^\mathcal{D}$}{sec:concepts:natural-indirect-effect}}

\tcontent{G}{D}{D}{\pref{Project}{sec:vocabulary:projection},\\ \pref{Rotate}{sec:vocabulary:rotation}}
\tcontent{H}{A}{B}{\pref{SP-LIME}{sec:ensemble:sp-lime}}
\tcontent{I}{A}{A}{\pref{Behavioral\\ Probes$^\mathcal{D}$}{sec:linguistic-information:behavioral-probes}}
\tcontent{I}{D}{D}{\pref{Structural \\ Probes$^\mathcal{D}$}{sec:linguistic-information:structural-probes}}
\tcontent{I}{E}{E}{\pref{Structural \\ Probes$^\mathcal{D}$}{sec:linguistic-information:structural-probes}}
\tcontent{I}{F}{F}{Auxiliary \\ Task$^\mathcal{D}$}
\tcontent{J}{A}{B}{\pref{SEAR$^\mathcal{M}$}{sec:rules:sear}}
\tcontent{J}{B}{E}{\pref{Compositional Explanations of Neurons$^{\dagger}$}{sec:rules:comp-explain-neuron}}

\end{tikzpicture}


%% file: chapters/motivation.tex
\section{Motivations for Interpretability}
\label{sec:motivation-for-interpretability}

The need for interpretability comes primarily from an ``incompleteness in the problem formalization'' \citep{Doshi-Velez2017a}, meaning if the model was constrained and optimized to prevent all possible ethical issues, interpretability would be much less relevant. However, because perfect optimization is unlikely, hence \motivation{safety} and \motivation{ethics} are strong motivations for interpretability.

Additionally, when models misbehave there is a need for explanations, to hold people or companies accountable, hence \motivation{acountability} is often a core motivation for interpretability. Finally, explanations are often useful, or sometimes necessary, for gaining \motivation{scientific understanding} about models. This section aims to elaborate on what exactly is understood by these terms and how interpretability can address them.

\paragraph{Ethics,} in the context of interpretability, is about ensuring that the model's behavior is aligned with common ethical and moral values. Because there does not exist an exact measure for this desideratum, this is ultimately something that should be judged qualitatively by humans, for example by an \emph{ethics review committee}, who will inspect the model explanations.

For some ethical concerns, such as discrimination, it may be possible to measure and satisfy this ethical concern via fairness metrics and debiasing techniques \citep{Garrido-Munoz2021}. However, this often requires a finite list of protected attributes \citep{Ho2021}, and such a list will likely be incomplete, hence the need for a qualitative assessment \citep{Doshi-Velez2017a, Lipton2018}.

\paragraph{Safety,} is about ensuring the model performs within expectations in deployment. As it is nearly impossible to truly test the model, in the end-to-end context that it will be deployed, ensuring \motivation{safety} does to some extent involve qualitative assessment \citep{Doshi-Velez2017a}. \citet{Lipton2018} frames this as \motivation{trust}, and suggests one interpretation of this is ``that we feel comfortable relinquishing control to it''.

While all types of interpretability can help with \motivation{safety}, in particular, \type{sec:adversarial-examples}{adversarial examples} and \type{sec:counterfactuals}{counterfactuals} are useful, as they evaluate the model on data outside the test distribution. \citet{Lipton2018} frames this in the broader context of \motivation{transferability}, which is the model's robustness to adversarial attacks and distributional shifts.

\paragraph{Accountability,} relates to explaining the model when it does fail in production. The ``right to explanation'', regarding the logic involved in the model's prediction, is increasingly being adopted, most notably in the EU via its GDPR legislation. However, also the US and UK have expressed support for such regulation \citep{Doshi-Velez2017}. Additionally, industries such as banking, are already required to audit their models \citep{Bhatt2020}.

\motivation{Accountability} is perhaps the core motivation of interpretability, as \citet{Miller2019} writes ``Interpretability is the degree to which a human can understand the cause of a decision'', and it is exactly the causal reasoning that is relevant in \motivation{accountability} \citep{Doshi-Velez2017}. 

\paragraph{Scientific Understanding,} addresses a need from researchers and scientists, which is to generate hypotheses and knowledge. As \citet{Doshi-Velez2017a} frames it, sometimes the best way to start such a process is to ask for explanations. In model development, explanations can also be useful for \motivation{model debugging} \citep{Bhatt2020}, which is often facilitated by the same kinds of explanations.

%% file: chapters/motivating_example.tex
\section{Motivating Example}
\label{sec:motivating-example}

Because \posthoc{post-hoc} methods are often model-agnostic, explaining and discussing them can often become abstract. To make the method sections as concrete and comparable as possible this survey will show fictive examples often based on the ``Stanford Sentiment Treebank'' (SST) dataset \citep{Socher2013}.
The SST dataset has been modeled using LSTM \citep{Wang2019}, Self-Attention-based models \citep{Devlin2019}, etc., all of which are popular examples of neural networks.

We use a sequence-to-class problem because this is what most interpretability methods applies to. Although some are agnostic to the problem type and others are specific to sequence-to-sequence problems. Throughout this survey we attempt to highlight what problems each method applies to. 

\begin{figure}[th]
    \centering
    \examplefigure{base}
    \caption{Three examples from the SST dataset \citep{Socher2013}. $\mathbf{x}$ is the input, with each token denoted by an \underline{underline}. $y$ is the gold target label, where \texttt{pos} is \emph{positive} and \texttt{neg} is \emph{negative} sentiment. Finally, $p(y|\mathbf{x})$ is the model's estimate of $\mathbf{x}$ belonging to category $y$. Note that the model predicts the 3rd (last) wrong, indicated with \textcolor{rgb,255:red,179; green,0; blue,0}{red}.}
    \label{fig:motivating_example:examples}
\end{figure}

The model responsible for the predictions in \Cref{fig:motivating_example:examples} can be explained by asking different questions, each of which communicates a different aspect of the model that is covered in the sections of this survey. Sometimes these explanation relates to a single observation, other times the explanation relates to the whole model.

\paragraph{local explanations} explain a single observation:
\begin{description}[align=right,labelwidth=9.5em,font={\normalfont\sffamily}]
    \item[Input Features] \emph{Which tokens are most important for the prediction, \Cref{sec:input-features}.}
    \item[Adversarial Examples] \emph{What would break the model's prediction, \Cref{sec:adversarial-examples}.}
    \item[Influential Examples] \emph{What training examples influenced the prediction, \Cref{sec:influential-examples}.}
    \item[Counterfactuals] \emph{What does the model consider a valid opposite example, \Cref{sec:counterfactuals}.}
    \item[Natural Language] \emph{What would a generated natural language explanation be, \Cref{sec:natural-language}.}
\end{description}

\paragraph{Class explanations} summarize the model, but only with regard to one selected class:
\begin{description}[align=right,labelwidth=9.5em,font={\normalfont\sffamily},itemsep=0.15em,parsep=0em,topsep=0.5em]
    \item[Concepts] \emph{What concepts (e.g. movie genre) can explain a class, \Cref{sec:concepts}.}
\end{description}

\paragraph{Global explanations} summarize the entire model with regards to a specific aspect:
\begin{description}[align=right,labelwidth=9.5em,font={\normalfont\sffamily},itemsep=0.15em,parsep=0em,topsep=0.5em]
    \item[Vocabulary] \emph{How does the model relate words to each other, \Cref{sec:vocabulary}.}
    \item[Ensemble] \emph{What examples are representative of the model, \Cref{sec:ensemble}.}
    \item[Linguistic information] \emph{What linguistic information does the model use, \Cref{sec:linguistic-information}.}
    \item[Rules] \emph{Which general rules can summarize an aspect of the model, \Cref{sec:rules}.}
\end{description}

%% file: chapters/measures.tex
\section{Measures of Interpretability}
\label{sec:measures-of-interpretability}

Because interpretability is by definition about explaining the model to humans \citep{Doshi-Velez2017a, Miller2019}, and these explanations are often qualitative, it is not clear how to quantitatively evaluate and compare interpretability methods. This ambiguity has lead to much discussion. Most notable is the \intrinsic{intrinsically} interpretable method \emph{Attention}, where different measures of interpretability have been published resulting in conflicting findings \citep{Jain2019,Serrano2019,Wiegreffe2020}. 

In general, there is no consensus on how to measure interpretability. However, validation is still paramount. As such, this section attempts to cover the general categories, themes, and methods that have been proposed. Additionally, each method section, starting from \type{sec:input-features}{input features}, in \Cref{sec:input-features}, will briefly cover how the authors choose to evaluate their method.

To describe the evaluation strategies, we use the terminology defined by \citet{Doshi-Velez2017a}, which separates the evaluation of interpretability into three categories, \measure{functionally-grounded}, \measure{human-grounded}, and \measure{application-grounded}. This categorization reflects the need to have explanations that are useful to humans (\measure{human-grounded}) and accurately reflect the model (\measure{functionally-grounded}). 

\paragraph{Application-grounded} evaluation is when the interpretability method is evaluated in the environment it will be deployed. For example, does the explanations result in higher survival-rates in a medical setting, a higher-grades in a homework-hint system, or a better model in a label-correction setting \citep{Doshi-Velez2017a, Williams2016}. Importantly, this evaluation should include the baseline where the explanations are provided by humans.

Due to the application-specific and time-consuming nature of this approach, \measure{application-grounded} evaluation is rarely done in NLP interpretability research. Instead, more synthetic and general evaluation setups can be used, which is what \measure{functionally-grounded} and \measure{human-grounded} evaluation is about. These categories each provide an important but different aspect for validating interpretability and should therefore be used in combination.

\paragraph{Human-grounded} evaluation checks if the explanations are useful to humans. Unlike \measure{application-grounded}, the task is often simpler and the task itself can be evaluated immediately. Additionally, expert humans are often not required \citep{Doshi-Velez2017a}. In other literature this is known as \measure{simulatability} \citep{Lipton2018} and \measure{comprehensibility} \citep{Robnik-Sikonja2018a}.

Although, \measure{human-grounded} evaluation is much more efficient than \measure{application-grounded} evaluation, the human aspect still takes time. An unfortunate but common approach is therefore to replace the human with a simulated user. This is unfortunate as providing explanations that are informative to humans is a non-trivial task, and often involves interdisciplinary knowledge from the human-computer interaction (HCI) and social science fields. Replacing a human with a simulated user, therefore leads to over optimistic results.

\citet{Miller2019} provides an excellent overview on what effective explanation is from the social science perspective, and criticizes current works by saying ``most work in explainable artificial intelligence uses only the researchers’ intuition of what constitutes a `good' explanation''.

It is therefore critical that interpretability methods are \measure{human-grounded}. These are common strategies to measure \measure{human-grounded}, used both in NLP and other fields:
\begin{itemize}
    \item Humans have to choose the best model based on an explanation \citep{Ribeiro2016}.
    \item Humans have to predict the model's behavior on new data \citep{Rajani2019}.
    \item Humans have to identify an outlier example called an intruder \citep{Chang2009}.  While it can be used on other fields, it is most common in NLP where it used with \type{sec:vocabulary}{vocabulary} explanations \citep{Park2017}.
\end{itemize}

\paragraph{Functionally-grounded} evaluation checks how well the explanation reflects the model. This is more commonly known as \measure{faithfulness} \citep{Ribeiro2016,Wiegreffe2020,Du2019,Jacovi2020} or sometimes \measure{fidelity} \citep{Robnik-Sikonja2018a}.

It might seem surprising that an explanation, which is directly produced from the model, would not reflect the model. However, even intrinsically interpretable methods such as \emph{Attention} and \emph{Neural Modular Networks} have been shown to not reflect the model \citep{Jain2019, Subramanian2020}.

Interestingly, \measure{human-grounded} interpretability methods can not reflect the model perfectly, because humans require explanations to be selective, meaning the explanation should select ``one or two causes from a sometimes infinite number of causes'' \citep{Miller2019}. Regardless, the explanations must still reflect the model to some extent, which surprisingly is not always the case \citep{Rudin2019, Jacovi2020}. Additionally, explanations that provide a similar type of explanation, with similar selectiveness, should compete on proving the explanation that best reflects the model.

For some tasks, measuring if an interpretability method is \measure{functionally-grounded} is trivial. In the case of \type{sec:adversarial-examples}{adversarial examples}, it is enough to show that the prediction changed and the adversarial example is a paraphrase. In other cases, most notably \type{sec:input-features}{input features}, providing a \measure{functionally-grounded} metric can be very challenging \citep{Jacovi2020,Kindermans2019,Yeh2019,Adebayo2018,Hooker2019}. 


In general, common evaluation strategies, both in NLP and other fields, are:
\begin{itemize}
    \item Comparing with an intrinsically interpretable model, such as logistic regression \citep{Ribeiro2016}.
    \item Comparing with other post-hoc methods \citep{Jain2019}.
    \item Proposing axiomatic desirables \citep{Sundararajan2017a}.
    \item Benchmarking against random explanations \citep{Hooker2019, Madsen2022}.
\end{itemize}

%% file: chapters/methods.tex
\section{Methods of Interpretability}

The main objective of this survey is to give an overview of \posthoc{post-hoc} interpretability methods and categorize them by how they communicate. \Cref{sec:input-features} to \Cref{sec:rules} will be dedicated towards this goal.

\Cref{tab:overview} represents a table-of-content, relating each section to a communication approach, but also contrasts the different methods by what information they use. In addition, the motivating example in \Cref{sec:motivating-example} gives a brief idea of the different communication approaches.

Each method section from \type{sec:input-features}{input features} (\cref{sec:input-features}) to \type{sec:rules}{rules} (\cref{sec:rules}) covers one communication approach, corresponding to one row in \Cref{tab:overview}, and can be read somewhat independently. Each section discusses the purpose of the communication approach and covers the most relevant methods and how they are evaluated. Because interpretability is a large field, this survey chooses methods based on historical progression and diversity regarding what information they use, this is discussed more in \hyperref[sec:limitations]{\emph{limitations}} (\cref{sec:limitations}). Finally, at the end of each method section we discuss the general trends and issues related to that communication approach.

Each method section will use the terminology\footnote{If you are viewing this paper in a PDF reader, each term will \emph{link} to the section where it's defined.} covered in \motivation{motivation for interpretability} (\cref{sec:motivation-for-interpretability}) and \measure{measures of interpretability} (\cref{sec:measures-of-interpretability}).


%% file: chapters/input_features.tex
\section{Input Features}
\label{sec:input-features}

An \type{sec:input-features}{Input feature} explanation is a \category{local explanation}, where the goal is to determine how important an \type{sec:input-features}{input feature}, e.g. a token, is for a given prediction. This approach is highly adaptable to different problems, as the input features are always known and are often meaningful to humans. Especially in NLP, the input features will often represent words, sub-words, or characters. Knowing which words are the most important, can be a powerful explanation method. An \type{sec:input-features}{input feature} explanation of the input $\mathbf{x}$, is represented as
\begin{equation}
    \mathbf{E}(\mathbf{x}, c): \mathrm{I}^\mathbf{d} \rightarrow \mathbb{R}^\mathbf{d} \text{, where $\mathrm{I}$ is the input domain and $\mathbf{d}$ is the input dimensionality.}
\end{equation}
Note that, when the output is a score of importance the explanation is called an \emph{importance measure}. 

Importantly, \type{sec:input-features}{input feature} explanations can only explain one scalar, meaning one class at one timestep. In a sequence-to-sequence application, the explanation is therefore repeated for each time step \citep{Li2016, Madsen2019a} although this may not respect the combinatorial complexities \citep{Alvarez-Melis2017}. Additionally, the selected class is either the most likely class or the true-label class, in this section the explained class is denoted with $c$. For all methods in this section, except \method{sec:input-features:anchors}{Anchors}, $c$ can be set as desired.

\input{chapters/input_features_gradient}
\ifarxiv{\input{chapters/input_features_ig}}
\input{chapters/input_features_lime}
\ifarxiv{\input{chapters/input_features_shap}}
\ifarxiv{\input{chapters/input_features_anchors}}

\subsection{Discussion}

\paragraph{Groundedness} The \measure{functionally-groundedness} of \type{sec:input-features}{input feature} explanations have recived a lot of attention and discussion, however there is still little consensus on what is \measure{functionally-grounded} or how to even measure it \cite{Bastings2021,Madsen2022,Vashishth2019,Jain2019,Serrano2019,Wiegreffe2020,Kindermans2019,Adebayo2018}.

\paragraph{Future work} It has been suggested, that a general \measure{functionally-grounded} post-hoc \type{sec:input-features}{input feature} explanation method just doesn't exists \citep{Rudin2019}, an analogue to the no-free-lunch theorem.
For this reason, a new trend in NLP is to develop architecture specific \type{sec:input-features}{input feature} explanations \citep{Brunner2020,Abnar2020a}, for example using attention. Although others are aganist this direction and do not think that attention can provide more \measure{functionally-grounded} explanations than general alternatives \citep{Bastings2020}.

Such high-level questions are likely difficult to answer without a more fundamental understanding of what the \measure{functionally-groundedness} desirables are. We therefore advocate for continuing the effort in measuring \measure{functionally-groundedness} but to focus more on establishing the fundamental desirables.


%% file: chapters/input_features_gradient.tex
\subsection{Gradient}
\label{sec:input-features:gradient}

One simple \type{sec:input-features}{importance measure}, is taking the gradient w.r.t. the input \citep{Baehrens2010, Li2016}.
\begin{equation}
\begin{aligned}
    \mathbf{E}_{\operatorname{gradient}}(\mathbf{x}, c) = L_p(\nabla_\mathbf{x} p(c|\mathbf{x};\theta)), &\text{ where } L_p \in \{L_1, L_2, L_\infty\} \\
    &\text{ and $p(c|\mathbf{x};\theta)$ is the model's probability output.}
\end{aligned}
\end{equation}

This essentially measures the change of the output, given an $\epsilon$-change to each input feature. Note, while NLP features are often discrete, it is still possible to take the gradient w.r.t. the one-hot-encoding by treating it as continuous. Although, because the one-hot-encoding has shape $\mathbf{x} \in \mathrm{I}^{V \times T}$, where $V$ is the vocabulary size and $T$ is the input length, it is necessary to reduce away the vocabulary dimension (often using an $L_p$-norm) such $\mathbf{E}(\mathbf{x}) \in \mathbb{R}^{T}$, when visualizing the importance per word as seen in \Cref{fig:input-features:gradient}.

The primary argument for the \method{sec:input-features:gradient}{gradient} method being \measure{functionally-grounded}, is that for a linear model $\mathbf{x}\mathbf{W}$, the explanation would be $\mathbf{W}^\top_{c,:}$ which is clearly a valid explanation \citep{Adebayo2018}. However, this does not guarantee  \measure{functionally-groundedness} for non-linear models, as the explanation mearly relates to a first-order Taylor approximation \citep{Li2016}. Additionally, areas of the input may be important but have zero gradients, this issue is discussed \Cref{sec:input-features:integrated-gradient}. 

Finally, it can be sensible to consier the scale of $\mathbf{x}$ too, hence the extension $\mathbf{x} \odot \nabla_\mathbf{x} p(c|\mathbf{x};\theta)$ is sometimes preferred. Although, a counter-argument is that $\mathbf{x}$ does not directly relate to the model, and this can therefore result in a less faithful explanation \citep{Adebayo2018}.

\begin{figure}[H]
    \centering
    \examplefigure{gradient}
    \caption{Hypothetical visualization of applying $\mathbf{E}_{\operatorname{gradient}}(\mathbf{x})$, where $c$ is the explained class. Note that because the vocabulary dimension is reduced away, typically using the $L^2$-norm, it is not possible to separate positive influence (red) from negative influence (blue).}
    \label{fig:input-features:gradient}
\end{figure}

%% file: chapters/input_features_ig.tex
\subsection{Integrated Gradient (IG)}
\label{sec:input-features:integrated-gradient}
The \method{sec:input-features:gradient}{gradient} approach has been further developed, the most notable development is \method{sec:input-features:integrated-gradient}{Integrated Gradient} \citep{Sundararajan2017a}.

\citet{Sundararajan2017a} primarily motivate \method{sec:input-features:integrated-gradient}{Integrated Gradient} via the desirables they call \emph{sensitivity} and \emph{completeness}. \emph{Sensitivity} means, if there exists a combination of $\mathbf{x}$ and baseline $\mathbf{b}$ (often an empty sequence), where the logit outputs of $f(\mathbf{x};\theta)$ and $f(\mathbf{b};\theta)$ are different, then the feature that changed should get a non-zero attribution. This desirable is not satisfied for the gradient method, for example due to the truncation  in $\operatorname{ReLU}(\cdot)$. \emph{Completeness} means, the sum of importance scores assigned to each token should equal the model output relative to the baseline $\mathbf{b}$.

To satify these desirables, \citet{Sundararajan2017a} develop equation \eqref{eq:input-features:integrated-gradient:formulation} which integrates the gradients between an uninformative baseline $\mathbf{b}$ and the observation $\mathbf{x}$ \citep{Sundararajan2017a}.
\begin{equation}
\begin{aligned}
    \mathbf{E}_{\operatorname{integrated-gradient}}(\mathbf{x}, c) &= (\mathbf{x} - \mathbf{b}) \odot \frac{1}{k} \sum_{i=1}^{k} \nabla_{\tilde{\mathbf{x}}_i} f(\tilde{\mathbf{x}}_i;\theta)_c, \quad \tilde{\mathbf{x}}_i = \mathbf{b} + \sfrac{i}{k}(\mathbf{x} - \mathbf{b}), \\
    \text{where }&\text{$f(\mathbf{x};\theta)$ is the model logits.}
    \label{eq:input-features:integrated-gradient:formulation}
\end{aligned}
\end{equation}

This approach has been successfully applied to NLP, where the uninformative baseline can be an empty sentence, such as padding tokens \citep{Mudrakarta2018}.

Although Integrated Gradient has become a popular approach, it has recently received criticism in computer vision (CV) community for not being \measure{functionally-grounded} \citep{Hooker2019}. More recent work have applied a similar analysis to NLP, and found that the \measure{functionally-groundedness} is at the very least task dependent \citep{Madsen2022}. Additonally, \citet{Bastings2021} uses synthetic NLP tasks and arrived at the same task-dependent conclusion. One explanation for the lack of \measure{functionally-groundedness} is the input mutiplication which is not directly related to the model \citep{Adebayo2018}.

%% file: chapters/input_features_lime.tex
\subsection{LIME}
\label{sec:input-features:lime}
Another popular approach is \method{sec:input-features:lime}{LIME} \citep{Ribeiro2016}. This distinguishes itself from the gradient-based methods by not relying on gradients. Instead, it samples nearby observations $\tilde{\mathbf{x}}$ and uses the model estimate $p(c|\tilde{\mathbf{x}})$ to fit a logistic regression. The parameters $\mathbf{w}$ of the logistic regression then represents the \type{sec:input-features}{importance measure}, since larger parameters would mean a greater effect on the output.
\begin{equation}
\begin{aligned}
    \mathbf{E}_{\operatorname{LIME}}(\mathbf{x}, c) = &\argmin_{\mathbf{w}} 
    \frac{1}{k} \sum_{i=1}^k \left(p(c|\tilde{\mathbf{x}}_i;\theta) \log(q(\tilde{\mathbf{x}}_i)) + (1-p(c|\tilde{\mathbf{x}}_i;\theta)) \log(1-q(\tilde{\mathbf{x}}_i)\right) + \lambda \|\mathbf{w}\|_1 \\
    &\text{ where } q(\tilde{\mathbf{x}}) = \sigma(\mathbf{w} \tilde{\mathbf{x}})
\end{aligned}
\end{equation}

One major complication of \method{sec:input-features:lime}{LIME} is how to sample $\tilde{\mathbf{x}}$, representing the nearby observations. In the original paper \citep{Ribeiro2016}, they use a Bag-Of-Words (BoW) representation with a cosine distance. While this approach remains possible with a model that works on sequential data, such distance metrics may not effectively match the model's internal space. In more recent work \citep{Wu2021}, they sample $\tilde{\mathbf{x}}$ by masking words of $\mathbf{x}$. However, this requires a model that supports such masking.

\begin{figure}[h]
    \centering
    \examplefigure{lime}
    \caption{A fictive visualization of LIME, where the weights of the logistic regression determine the \type{sec:input-features}{importance measure}. Note that for LIME, it is possible to have negative importance (indicated by blue). Furthermore, some tokens have no importance score, due to the $L^1$-regularizer.}
    \label{fig:input-features:lime}
\end{figure}

The advantages of \method{sec:input-features:lime}{LIME} are that it only depends on black-box information and the dataset, therefore no gradient calculations are required. Secondly, it uses a LASSO logistic regression, which is a normal logistic regression with an $L_1$-regularizer. This means that its explanation is selective, as in sparse, which may be essential for providing a human-friendly explanation \citep{Miller2019}. 

\citet{Ribeiro2016} show that LIME is \measure{functionally-grounded} by applying LIME on \intrinsic{intrinsically} interpretable models, such as a logistic regression model, and then compare the LIME explanation with the \intrinsic{intrinsic} explanation from the logistic regression. They also show \measure{human-groundedness} by conducting a human trial experiment, where non-experts have to choose the best model, based on the provided explanation, given a ``wrong classifer'' tranined on a bias dataset and a ``correct classifer'' trained on a curated dataset.

%% file: chapters/input_features_shap.tex
\subsection{Kernel SHAP}
\label{sec:input-features:shap}

A limitation of \method{sec:input-features:lime}{LIME} is that the weights in a linear model are not necessarily \intrinsic{intrinsically} interpretable. When there exists multicollinearity (input features are linearly correlated with each other) then the model weights can be scaled arbitrarily creating a false sense of importance.

To avoid the multicollinearity issue, one approach is to compute Shapley values \citep{Shapley1953} which are derived from game theory. The central idea is to fit a linear model for every permutation of features enabled. For example, if there are two features $\{x_1, x_2\}$, the Shapley values would aggregate the weights from fitting the datasets with features $\{\varnothing\}, \{x_1\}, \{x_2\}, \{x_1, x_2\}$. If there are $T$ features this would require $\mathcal{O}(2^T)$ models.

While this method works in theory, it is clearly intractable. \citet{Lundberg2017} present a framework for producing Shapley values in a more tractable manner. The model-agnostic approach they introduce is called \method{sec:input-features:shap}{Kernel SHAP}. It combines 3 ideas: it reduces the number of features via a mapping function $h_\mathbf{x}(\mathbf{z})$, it uses squared-loss instead of cross-entropy by working on logits, and it weighs each observation by how many features there are enabled.

\begin{equation}
\begin{aligned}
    \mathbf{E}_{\operatorname{SHAP}}(\mathbf{x}, c) = &\argmin_{\mathbf{w}} 
    \sum_{\mathbf{z} \in \mathbb{Z}^M} \pi(\mathbf{z})\ (f(h_\mathbf{x}(\mathbf{z});\theta)_c - g(\mathbf{z}))^2 \\
    &\text{where } g(\mathbf{z}) = \mathbf{w} \mathbf{z} \\
    &\phantom{\text{where }} \pi(\mathbf{z}) = \frac{M - 1}{(M\, \operatorname{choose}\, |\mathbf{z}|) |\mathbf{z}| (M - |\mathbf{z}|)}
\end{aligned}
\label{eq:input-features:shap}
\end{equation}

In \eqref{eq:input-features:shap}, $\mathbf{z}$ is a $\{0,1\}^M$ vector that describes which combined features are enabled. This is then used in $h_\mathbf{x}(\mathbf{z})$, which enables those features in $\mathbf{x}$. Furthermore, $\mathbb{Z}^M$ represents all permutations of enabled combined features and $|\mathbf{z}|$ is the number of enabled combined-features. \Cref{fig:input-features:shap}, demonstrates a fictive example of how input features can be combined and visualize their shapley values.

\begin{figure}[h]
    \centering
    \examplefigure{shap}
    \caption{Fictive visualization of \method{sec:input-features:shap}{Kernel SHAP}. Note how input tokens are combined to a single feature to make \method{sec:input-features:shap}{SHAP} more tractable to compute, this is the role of $h_\mathbf{x}(z)$ in \eqref{eq:input-features:shap}.}
    \label{fig:input-features:shap}
\end{figure}

\citet{Lundberg2017} show \measure{functionally-groundedness} by using that Shapley values uniquely satisfy a set of desirables and that \method{sec:input-features:shap}{SHAP} values are also Shapley values. Furthermore, \citet{Lundberg2017} show \measure{human-groundedness} by asking humans to manually produce importance measures and correlate them with the \method{sec:input-features:shap}{SHAP} values.

A criticism of both SHAP and LIME is that they depend on pertubation of the input, this makes it possible to create adverserial models that appear ethical when explained using pertubated inputs but is in reality not ethical when evaluted without pertubation \citep{Slack2020a}. This means that LIME and SHAP can only provide a \measure{functionally-grounded} explanation as long as the model is trained without malicious intent.

\method{sec:input-features:shap}{SHAP} and Shapley values in general are heavily used in the industry \citep{Bhatt2020}. In NLP literature \method{sec:input-features:shap}{SHAP} has been used by \citet{Wu2021}. This popularity is likely due to their mathematical foundation and the \texttt{shap} library. In particular, the \texttt{shap} library also presents Partition SHAP which claims to reduce the number of model evaluations to $M^2$, instead of $2^M$ \footnote{See documentation \url{https://shap.readthedocs.io/en/latest/example_notebooks/tabular\_examples/model_agnostic/Simple\%20Boston\%20Demo.html}}. One major disadvantage of \method{sec:input-features:shap}{SHAP} is it inherently depends on the masked inputs still being valid inputs. For some NLP models, this can be accomplished with a \texttt{[MASK]} token, while for it is not possible in a \posthoc{post-hoc} setting. For this reason, \method{sec:input-features:shap}{SHAP} exists at an intersection between \posthoc{post-hoc} and \posthoc{intrinsic} interpretability methods. This intersection is discussed more in \Cref{sec:future-directions}.





%% file: chapters/input_features_anchors.tex
\subsection{Anchors}
\label{sec:input-features:anchors}

A further development of the idea, that sparse explanations are easier to understand, is \method{sec:input-features:anchors}{Anchors}. Instead of giving an importance score, like in the case of the gradient-based methods or \method{sec:input-features:lime}{LIME}, the \method{sec:input-features:anchors}{Anchors} simply provides a shortlist of words that were most relevant for making the prediction \citep{Ribeiro2018a}. The authors show \measure{human-groundedness} with a similar user setup as in \method{sec:input-features:lime}{LIME} \citep{Ribeiro2016}.

\begin{figure}[h]
    \centering
    \examplefigure{anchor}
    \caption{Fictive visalization, showing the \method{sec:input-features:anchors}{anchors} that are responsible for the prediction.}
    \label{fig:input-features:anchors}
\end{figure}

The list-of-words called ``anchors'' ($A$) is formalized in \eqref{eq:input-features:anchors}. Note that $c = \argmax_i p(i|\mathbf{x};\theta)$ is a requirement for \method{sec:input-features:anchors}{anchors}, as using $\operatorname{prec}(A) = \mathbb{E}_{\mathcal{D}(\tilde{\mathbf{x}}|A)} \left[\mathds{1}_{y = \tilde{y}}\right]$ in \eqref{eq:input-features:anchors} would cause \method{sec:input-features:anchors}{anchors} to be unaffected by the model.

\begin{equation}
\begin{aligned}
    \mathbf{E}_{\operatorname{anchors}}(\mathbf{x}) = &\argmax_{A \text{ s.t. } \operatorname{prec}(A) \ge \tau\, \wedge\, A(\mathbf{x}) = 1} \operatorname{cov}(A) \\
    &\text{where } \operatorname{prec}(A) = \mathbb{E}_{\mathcal{D}(\tilde{\mathbf{x}}|A)} \left[\mathds{1}_{[\argmax_i p(i|\mathbf{x};\theta) = \argmax_i p(i|\tilde{\mathbf{x}};\theta)]}\right] \\
    &\phantom{\text{where }} \operatorname{cov}(A) = \mathbb{E}_{\mathcal{D}(\tilde{\mathbf{x}})} \left[A(\tilde{\mathbf{x}})\right] \\
    &\phantom{\text{where }} A(\mathbf{x}) = \begin{cases}
    1 & \text{if the anchors $A$ are in $\mathbf{x}$} \\
    0 & \text{otherwise}
    \end{cases}
\end{aligned}
\label{eq:input-features:anchors}
\end{equation}
This formalization says the anchor words should have the highest coverage ($\operatorname{cov}(A)$), meaning the most sentences in the dataset $D(\tilde{\mathbf{x}})$ contains the anchors $A$. Furthermore, only consider anchors $A$ that are sufficiently precise ($\operatorname{prec}(A) \ge \tau$) and in $\mathbf{x}$. Precision is defined as the ratio of observations $\tilde{\mathbf{x}}$ with anchors $A$, denoted $\mathcal{D}(\tilde{\mathbf{x}}|A)$, where the predicted label of $\tilde{\mathbf{x}}$ matches the predicted label of $\mathbf{x}$.

Solving this optimization problem exactly is infeasible, as the number of anchors is combinatorially large. To approximate it, \citet{Ribeiro2018a} model $\operatorname{prec}(A) \ge \tau$ probabilistically \citep{Kaufmann2013} and then use a bottom-up approach, where they add a new word to the $k$-best anchor candidate in each iteration similar to beam-search.

%% file: chapters/adversarial_examples.tex
\section{Adversarial Examples}
\label{sec:adversarial-examples}

An \type{sec:adversarial-examples}{adversarial example}, is an input that causes a model to produce a wrong prediction, due to limitations of the model. The adversarial example is often produced from an existing example, for which the model produces a correct prediction. Because the \type{sec:adversarial-examples}{adversarial example} serves as an explanation, in the context of an existing example it is a \category{local explanation}.

\citet{Wang2019a} provide a thorough survey on \type{sec:adversarial-examples}{adversarial example} explanations, and also goes in depth regarding taxonomy, using \type{sec:adversarial-examples}{adversarial examples} for robustness, and similarity scores between the existing example and the \type{sec:adversarial-examples}{adversarial example}. Additonally, the survey by \citet{Belinkov2019} also have a section on adversarial examples.

In this survey we therefore focus on just two explanation methods. These \type{sec:adversarial-examples}{adversarial example} methods informs us about the support boundaries of a given example, which then informs us about the logic involved and therefore provides interpretability. In fact, this explanation can be similar to the \type{sec:input-features}{input feature} methods, discussed in \Cref{sec:input-features}. Many of those methods also indicate what words should be changed to alter the prediction. An important difference is that \type{sec:adversarial-examples}{adversarial} explanations are contrastive, meaning they explain by comparing with another example, while \type{sec:input-features}{input features} explain only concerning the original example. Contrastive explanations are, from a social science perspective, generally considered more \measure{human-grounded} \citep{Miller2019}.

In the following discussions, we refer the original example as $\mathbf{x}$ and the adversarial example as $\tilde{\mathbf{x}}$. The goal is to develop an adversarial method $A$, that maps from $\mathbf{x}$ to $\tilde{\mathbf{x}}$:
\begin{equation}
    A(\mathbf{x}) \rightarrow \tilde{\mathbf{x}}
\end{equation}

Importantly, to ensure that an \type{sec:adversarial-examples}{adverserial example} method is \measure{functionally-grounded}, one only needs to assert that the predicted label changes while the gold label remains the same. Additionally, it's a desireable to have the original and adverserial example to be similar, in many applications this can be framed as paraphrasing. Compared to other explanation types, these properties are reasonably trivial to measure. See \Cref{sec:measures-of-interpretability} for a general discussion on measures of interpretability.

Finally, because \type{sec:adversarial-examples}{adverserial example} explanations are framed by the output class, these explanations do not generalize easily to sequence-to-sequence problems. Although one could imagine for example an offensive-text classifier, which reduces the sequence-to-sequence model back to a sequence-to-class model.

\input{chapters/adversarial_examples_hotflip}
\ifarxiv{\input{chapters/adversarial_examples_sea}}

\subsection{Discussion}

\paragraph{Groundedness} \type{sec:adversarial-examples}{Adversarial example} are as mentioned, easy to measure \measure{functionally-groundedness} on and should be \measure{human-grounded} due to their contrastive nature \citep{Miller2019}. However, we are not aware of any work which explicitly tests for \measure{human-groundedness}. This is likely because it is considered to be a given, but we advocate for testing such a hypothesis anyway.

\paragraph{Future work} The difficulty with \type{sec:adversarial-examples}{adversarial example} explanations lies in the search procedure. For example, \method{sec:adversarial-examples:hotflip}{HotFlip} \citep{Ebrahimi2018} uses a greedy sequential search algorithm and would therefore not be able to identify combinatorial effects like a double-negative. While \method{sec:adversarial-examples:sea}{SEA} \citet{Ribeiro2018} depends on an expensive paraphrase generation model.

One typical limitation of \type{sec:adversarial-examples}{adversarial example} methods is that they provide no control of the search direction. Hypothetically, while changing ``unpredictable'' to ``unforeseeable'' could provide the largest source of error due to a robustness issue, it might be more interesting to discover that changing ``womans' chess club'' to ``mens' chess club'' also flips the label. Unfortunately, this aspect is usually not considered because the motivation for \type{sec:adversarial-examples}{adversarial example} generation is often robustness and debiasing.

%% file: chapters/adversarial_examples_hotflip.tex
\subsection{HotFlip}
\label{sec:adversarial-examples:hotflip}
A great example of the relation between \type{sec:input-features}{input feature} explanations and \type{sec:adversarial-examples}{adversarial examples} is \method{sec:adversarial-examples:hotflip}{HotFlip} \citep{Ebrahimi2018}. Here the effect of changing token $v$ to another token $\tilde{v}$ at position $t$, on the model loss $\mathcal{L}$, is estimated via using gradients
\begin{equation}
    \mathcal{L}(y, \tilde{\mathbf{x}}_{t: v\rightarrow \tilde{v}}) - \mathcal{L}(y, \mathbf{x};\theta) \approx \frac{\partial \mathcal{L}(y, \mathbf{x};\theta)}{\partial x_{t,\tilde{v}}} - \frac{\partial \mathcal{L}(y, \mathbf{x};\theta)}{\partial x_{t,v}},
\end{equation}
where $\tilde{\mathbf{x}}_{t: v\rightarrow \tilde{v}}$ is the one-hot-encoded input $\mathbf{x}$, with the token $v$ at position $t$ changed to $\tilde{v}$. Additionally, $x_{t,\tilde{v}}$ and $x_{t,v}$ are the scalar components of the one-hot-encoded input $\mathbf{x}$. 

Had a gradient approximation not been used, the alternative would be to exactly compute a forward pass for every possible token swap. Instead, this approximation only requires one backward pass. To produce an adversarial sentence with multiple tokens changed, the authors use a beam-search approach. A visualization of \method{sec:adversarial-examples:hotflip}{HotFlip} can be seen in  \Cref{fig:adversarial-examples:hotflip}.
\begin{equation}
    A_{\operatorname{HotFlip}}(\mathbf{x}) = \argmax_{\tilde{\mathbf{x}}_{t: v\rightarrow \tilde{v}}} \frac{\partial \mathcal{L}(y, \mathbf{x};\theta)}{\partial x_{t,\tilde{v}}} - \frac{\partial \mathcal{L}(y, \mathbf{x};\theta)}{\partial x_{t,v}}
\end{equation}

\begin{figure}[H]
    \centering
    \examplefigure{hotflip}
    \caption{Hypothetical visualization of \method{sec:adversarial-examples:hotflip}{HotFlip}. The highlight indicates the gradient w.r.t. the input, which HotFlip uses to select which token to change. $\mathbf{x}$ indicates the original sentence, and $\tilde{\mathbf{x}}$ indicates the adversarial sentence.}
    \label{fig:adversarial-examples:hotflip}
\end{figure}

The \method{sec:adversarial-examples:hotflip}{HotFlip} paper \citep{Ebrahimi2018} primarily investigates character-level models, for which the desire is to build a model that is robust against typos. However, in terms of word-level models, it is necessary to constrain the possible changes, such that the adversarial sentence is a paraphrase. They do this via the word-embeddings, such that the adversarial word and the original word are constrained to have a cosine similarity of at least 0.8.

The \method{sec:adversarial-examples:hotflip}{HotFlip} approach has proven effective for other adversarial explanation methods, such as the aforementioned Universal Adversarial Triggers \citep{Wallace2020}. 

%% file: chapters/adversarial_examples_sea.tex
\subsection{Semantically Equivalent Adversaries (SEA)}
\label{sec:adversarial-examples:sea}
An alternative approach to produce adversarial examples that are ensured to be paraphrases is to sample from a paraphrasing model $q(\tilde{\mathbf{x}} | \mathbf{x})$. \citet{Ribeiro2018} do this by measuring a semantical-equivalency-score $S(\mathbf{x}, \tilde{\mathbf{x}})$, as the relative likelihood of $q(\tilde{\mathbf{x}} | \mathbf{x})$ compared to $q(\mathbf{x} | \mathbf{x})$. It is then possible to maximize the similarity, while still having a different model prediction. The exact method is defined in \eqref{eq:adversarial-examples:sea}, which also constrains the optimization with a minimum semantical-equivalency-score and ensures the predicted label is different.

\begin{equation}
\begin{aligned}
    A_{\operatorname{SEA}}(\mathbf{x}) = \argmax_{\tilde{\mathbf{x}} \sim q(\tilde{\mathbf{x}} | \mathbf{x})}\ & S(\mathbf{x}, \tilde{\mathbf{x}}) \\
    \text{s.t. }& S(\mathbf{x}, \tilde{\mathbf{x}}) \ge 0.8 \\
    \phantom{s.t. }& \argmax_i p(i|\mathbf{x};\theta) \not= \argmax_i p(i|\tilde{\mathbf{x};\theta}) \\
    \text{where }& S(\mathbf{x}, \tilde{\mathbf{x}}) = \min\left(1, \frac{q(\tilde{\mathbf{x}} | \mathbf{x})}{q(\mathbf{x} | \mathbf{x})}\right)
\end{aligned}
\label{eq:adversarial-examples:sea}
\end{equation}

The reason why a relative score is necessary, as opposed to just using $S(\mathbf{x}, \tilde{\mathbf{x}}) = q(\tilde{\mathbf{x}}|\mathbf{x})$, is that for two normal sentences $\mathbf{x}_1$ and $\mathbf{x}_2$ of different length, longer sentences are just inherently less likely. Therefore, to maintain a comparative semantical-equivalency-score normalizing by $q(\mathbf{x}|\mathbf{x})$ is necessary \citep{Ribeiro2018}.

\begin{figure}[h]
    \centering
    \examplefigure{sea}
    \caption{Hypothetical results of using \method{sec:adversarial-examples:sea}{SEA} \citep{Ribeiro2018}. Note that unlike \method{sec:adversarial-examples:hotflip}{HotFlip}, \method{sec:adversarial-examples:sea}{SEA} can change and delete multiple tokens simultaneously as it samples from a paraphrasing model. Again, $\mathbf{x}$ indicates the original sentence, $\tilde{\mathbf{x}}$ indicates the adversarial sentence, and $S(\mathbf{x}, \tilde{\mathbf{x}})$ is the semantical-equivalency-score which must be at least $0.8$.}
    \label{fig:adversarial-examples:sea}
\end{figure}

%% file: chapters/influential_examples.tex
\section{Similar examples}
\label{sec:influential-examples}

For a given \emph{input example}, an \type{sec:influential-examples}{influential examples} explanation finds examples from the training dataset, that in terms of the model's understanding, looks like the \emph{input example}. Because this explanation method centers around a specific \emph{input example} it is a \category{local explanation}. Note that is different from just an distance metric on the inputs, such as BLEU \citep{Papineni2001}, as this does not depend on the model.

\type{sec:influential-examples}{Influential examples} explanations can be quite useful. For example, for discovering dataset artifacts as some of the \emph{influential examples} may have nothing to do with the \emph{input example}, except for the artifacts. Additonally, they are commonly used to discover mislabeled observations.

The \type{sec:influential-examples}{influential examples} can always be presented as just the examples and a similarity score, see \Cref{fig:influential-examples}. Because the only presentation difference is the similarity score, this chapter does not include example figures for each method.

\begin{figure}[h]
    \centering
    \examplefigure{influence}
    \caption{Fictive result showing the \emph{influential examples} $\tilde{\mathbf{x}}$, in relation to the \emph{input example} $\mathbf{x}$, showing both examples with positive and negative influence. $\Delta$ is the similarity score, the scale and range may depend on the specific method. Note, it is possible to measure the influence of an example on itself. This can be useful to identify mislabled observations, as such observations will be important for their own prediction.}
    \label{fig:influential-examples}
\end{figure}


\input{chapters/influential_examples_if}
\input{chapters/influential_examples_rps}
\input{chapters/influential_examples_tracin}

\subsection{Discussion}

\paragraph{Groundedness} \type{sec:influential-examples}{Influential example} explanations, is one of the few categories with a non-trival but appropiate \measure{functionally-grounded} metric, namely the label-correction experiment, which is used somewhat consistently across papers. Unfortunately, this experiment has not been used on NLP tasks and in general very little \measure{functionally-grounded} validation have been done in NLP. 

Additionally, the label-correction experiment is somewhat limited, as it evaluates the influence of a training observation on itself. This is not how a \type{sec:influential-examples}{Influential examples} explanation would be used in most applications, for example dataset artifact discovery. We therefore suggest future work also include the experiment from \citet{Guo2020} which uses information removal.

\paragraph{Future work} A natural question, when asking what training observations are influencial is to also what part of them are important. \method{sec:influential-examples:influence-functions}{Influence functions} can answer this, although at an increased computational cost. However \method{sec:influencial-examples:tracin}{TracIn} can not. For sequential outputs it is interesting to also be able to select parts of the output and ask what influenced this. Both of these questions, are becomming increasingly relevant with large-scale langauge models, where there is a large interest in understand what caused a particular generation.

%% file: chapters/influential_examples_if.tex
\subsection{Influence functions}
\label{sec:influential-examples:influence-functions}

\method{sec:influential-examples:influence-functions}{Influence functions} is a classical technique from robust statistics \citep{Cook1980}. However, in robust statistics, there are strong assumptions regarding convexity, low-dimensionality, and differentiability. Recent efforts in deep learning remove the low-dimensionality constraint and to some extent the convexity constraint \citep{Koh2017}.

The central idea in \method{sec:influential-examples:influence-functions}{influence functions}, is to estimate the effect on the loss  $\mathcal{L}$, of removing the observation $\tilde{\mathbf{x}}$ from the dataset. The most influential examples are those where the loss changes the most. Let $\tilde{\theta}$ be the model parameters if $\tilde{\mathbf{x}}$ had not been included in the training dataset, then the loss difference can be estimated using
\begin{equation}
\mathcal{L}(y, \mathbf{x}; \tilde{\theta}) - \mathcal{L}(y, \mathbf{x}; \theta) \approx \frac{1}{n} \nabla_{\theta} \mathcal{L}(y, \mathbf{x}; \theta)^{\top} H_{\theta}^{-1} \nabla_{\theta} \mathcal{L}(\tilde{y}, \tilde{\mathbf{x}}; \theta).
\label{eq:influential-examples:influence-functions:main}
\end{equation}

Importantly, the Hessian $H_{\theta}$ needs to be positive-definite, which can only be guaranteed for convex models. The authors \citet{Koh2017} avoid this issue, by adding a diagonal to the Hessian, until it is positive-definite. Additionally, they solve the computational issue of computing an inverse Hessian, by formulating \eqref{eq:influential-examples:influence-functions:main} as an inverse Hessian-vector product. Such formulation can be approximated in $\mathcal{O}(np)$ time, where $n$ is the number of observations and $p$ is the number of parameters, hence a computational complexity identical to one training epoch. Note however, that the inverse Hessian-vector product needs to be computed for every explained test observation $\mathbf{x}$.

One limitation of \method{sec:influential-examples:influence-functions}{influence functions} is that computing the \method{sec:influential-examples:influence-functions}{influence functions} is not always numerically stable \citep{Yeh2018}, because \eqref{eq:influential-examples:influence-functions:main} uses the gradient $\nabla_{\theta} \mathcal{L}(\tilde{y}, \tilde{\mathbf{x}}; \theta)$ which is optimized to be close to zero.


\citet{Koh2017} looked at support-vector-machines, which are known to be convex, and convolutional neural networks which are generally non-convex. \citet{Han2020} then extended the analysis of \method{sec:influential-examples:influence-functions}{influence functions} to BERT \citep{Devlin2019}. This is a crucial step, as BERT may be much further from convexity than CNNs, thus cause the \method{sec:influential-examples:influence-functions}{influence functions} to be less \measure{functionally-grounded}.

\citet{Han2020} validates for \measure{functionally-groundedness} by removing the 10\% most influential training examples from the dataset and then retrain the model. The results show a significant decrease in the model's performance on the test split, compared to removing the 10\% least influential examples and 10\% random examples, validating that the influential examples are important.

Additionally, \citet{Koh2017} measures \measure{functionally-groundedness} by setting 10\% of training observations to a wrong label. \method{sec:influential-examples:influence-functions}{Influence functions} is then used to select a fraction of the dataset, for which labels are corrected. The metric is then how many mislabeled observations were identified and the performance difference. The idea being, wrongly labeled observations should affect the loss more than correctly labeled observations, hence \method{sec:influential-examples:influence-functions}{influence functions} will tend to find wrongly labeled observations. \citet{Han2020} perform a similar experiment, but instead removes observations based on importance and then measures the performance difference. Both experiments validates that \method{sec:influential-examples:influence-functions}{influence functions} are \measure{functionally-grounded}.


\paragraph{Performance considerations.} A criticism of influence functions has been that it is computationally expensive. Although $\nabla_{\theta} \mathcal{L}(y, \mathbf{x}; \theta)^{\top} H_{\theta}^{-1}$ can be cached for each test example, it is still too computationally intensive for real-time inspection of the model. Additionally, having to compute the weight-gradient $\nabla_{\theta} \mathcal{L}(\tilde{y}, \tilde{\mathbf{x}}; \theta)$ and inner-product for every training observation, does not scale sufficiently. To this end, \citet{Guo2020} propose to only use a subset of training data, using a KNN clustering. Additionally, they show that the hyperparameters when computing $\nabla_{\theta} \mathcal{L}(y, \mathbf{x}; \theta)^{\top} H_{\theta}^{-1}$ can be tuned to reduce the computation to less than half.

%% file: chapters/influential_examples_rps.tex
\subsection{Representer Point Selection}
\label{sec:influential-examples:representer-point-selection}

An alternative to \method{sec:influential-examples:influence-functions}{influence functions}, is the Representer theorem \citep{Scholkopf2001}. The central idea is that the logits of a test example $\mathbf{x}$, can be expressed as a decomposition of all training samples $f(\mathbf{x}) = \sum_{i=1}^n \bm{\alpha}_i \kappa(\mathbf{x}, \tilde{\mathbf{x}}_i)$. The original Representer theorem \citep{Scholkopf2001} works on \emph{reproducing kernel Hilbert spaces}, which is not applicable for deep learning. However, recent work has applied the idea to neural networks \citep{Yeh2018}.

Let $\bm{\theta}_L$ be the weight matrix of the final layer, such that the logits $f(\mathbf{x};\theta) = \bm{\theta}_L \mathbf{z}_{L-1}(\mathbf{x};\theta)$, then if the regularized loss $\frac{1}{n} \sum_{i=1}^n \mathcal{L}(\tilde{y}_i, \tilde{\mathbf{x}}_i; \theta) + \lambda \|\bm{\theta}_L\|^2$, is a stationary point and $\lambda > 0$, then
\begin{equation}
    f(\mathbf{x}) = \sum_{i=1}^n \bm{\alpha}_i \mathbf{z}_{L-1}(\tilde{\mathbf{x}}_i;\theta)^\top \mathbf{z}_{L-1}(\mathbf{x};\theta), \text{ where } \bm{\alpha}_i = \frac{1}{2 \lambda \cdot n} \frac{\partial \mathcal{L}(\tilde{y}_i, \tilde{\mathbf{x}}_i; \theta)}{\partial \mathbf{z}_{L-1}(\mathbf{x}_i;\theta)} \ .
\end{equation}

To understand the importance of each training observation $\tilde{\mathbf{x}}_i$, regarding the prediction of class $c$ for the test example $\mathbf{x}$, one just looks at the $c$'th element of each term $\bm{\alpha}_i \mathbf{z}_{L-1}(\tilde{\mathbf{x}}_i;\theta)^\top \mathbf{z}_{L-1}(\mathbf{x};\theta)$. This approach is more numerically stable than \method{sec:influential-examples:influence-functions}{influence functions} \citep{Yeh2018}, but has the downside of only depending on intermediate representation of the final layer, while \method{sec:influential-examples:influence-functions}{influence functions} employs the entire model.

Because \method{sec:influential-examples:representer-point-selection}{Representer Point Selection} does depend on a specific model setup, where the last layer is regularized, this could be considered an \intrinsic{intrinsic} method. However, \citet{Yeh2018} show that the stationary solution can be achieved \intrinsic{post-hoc}, meaning after learning, with minimal impact on the model predictions. They do this via the optimization problem
\begin{equation}
\bm{\theta}_L = \argmin_{\bm{W}} \left( \frac{1}{n} \sum_{i=1}^n \mathcal{L}(p(\cdot|\tilde{\mathbf{x}}_i;\theta), \bm{W} \mathbf{z}_{L-1}(\tilde{\mathbf{x}}_i;\theta)) + \lambda \|\bm{W}\|^2\right),
\end{equation}
where $\theta$ is the original model parameters, $\bm{\theta}_L$ are the new parameters for the last layer, and $\mathcal{L}$ is the full cross-entropy loss. Because this is a fairly low-dimensional problem, fine-tuning this can be done with an L-BFGS optimizer or similar \citep{Yeh2018}.

\citet{Yeh2018} show this method is \measure{functionally-grounded} on a computer vision task, using a label-correction experiment similar to that in \method{sec:influential-examples:influence-functions}{influence functions}. In this case, $|\bm{\alpha}_{i,c}|$ is used to select the observations to perform label correction on. Their results show that \method{sec:influential-examples:representer-point-selection}{Representer Point Selection} and \method{sec:influential-examples:influence-functions}{influence functions} can identify wrong labels equally well, but that the observations which \method{sec:influential-examples:representer-point-selection}{Representer Point Selection} selects affects the models performance more. Unfortunately, \citet{Yeh2018} do only show anecdotal results on an NLP task.

%% file: chapters/influential_examples_tracin.tex
\subsection{TracIn}
\label{sec:influencial-examples:tracin}

The idea behind \method{sec:influencial-examples:tracin}{TracIn} by \citet{Garima2020} is to accumulate loss changes during training. Specifically, the loss change on the test observation $\mathbf{x}$ when optimizing $\tilde{\mathbf{x}}$. \citet{Garima2020} first introduce an idealized version of this, which assumes optimization is done on one observation at a time (for example, SGD):
\begin{equation}
\begin{aligned}
\operatorname{TracInIdeal}(\tilde{\mathbf{x}}, \mathbf{x}) = \sum_{t \in \mathcal{T}_{\tilde{\mathbf{x}}}} \mathcal{L}(y, \mathbf{x}, \theta_t) - \mathcal{L}(y, \mathbf{x}, \theta_{t+1}), \text{where } \mathcal{T}_{\tilde{\mathbf{x}}} \text{ is timestep which optimized } \tilde{\mathbf{x}}
\end{aligned}
\end{equation}

\method{sec:influencial-examples:tracin}{TracIn} TracIn is then a relaxation of this idealized version. Rather than using a direct loss difference, gradients are used. Rather than assuming stochastic gradient descent (or similar) mini-batches can be used. Rather than checking every time step, checkpoints collected during training can be used.

\begin{equation}
\begin{aligned}
\operatorname{TracIn}(\tilde{\mathbf{x}}, \mathbf{x}) &= \frac{1}{b} \sum_{t \in \mathcal{C}} \eta_t \nabla_{\theta_t} \mathcal{L}(y, \mathbf{x}, \theta_t) \cdot \nabla_{\theta_t} \mathcal{L}(\tilde{y}, \tilde{\mathbf{x}}, \theta_{t}), \\
\text{where }& \text{$\mathcal{C}$ are checkpoints, $b$ is batch-size, and $\eta_t$ is learning-rate.}
\end{aligned}
\label{eq:tracin:formulation}
\end{equation}

Note, that the \eqref{eq:tracin:formulation} formulation is still based on plain gradient descent. However, \citet{Garima2020} instruct how to adapt this to most learning algorithms (AdaGrad, Adam, Newton, etc).

As a \measure{functionally-grounded} evaluation, \citet{Garima2020} repeat the label-correction experiment of \method{sec:influential-examples:influence-functions}{influence functions} and \method{sec:influential-examples:representer-point-selection}{Representer Point Selection}, and find that their method can better select mislabeled observations. Note that this was evaluated on CIFAR-10 and MNIST. Unfortunately, \citet{Garima2020} does not do any evaluation on NLP tasks, but they do anecdotally show it works on an NLP application.

%% file: chapters/counterfactuals.tex
\section{Counterfactuals}
\label{sec:counterfactuals}

\type{sec:counterfactuals}{Counterfactual explanations} are essentially answering the question ``how would the input need to change for the prediction to be different?''. Furthermore, these \type{sec:counterfactuals}{counterfactual examples} should be a minimal-edit from the original example and fluent. However, all of these properties can also be said of \type{sec:adversarial-examples}{adversarial explanations}, and indeed some works confuse these terms. The critical difference is that \type{sec:adversarial-examples}{adversarial examples} should have the same gold label as the original example, while \type{sec:counterfactuals}{counterfactual examples} should have a different gold label (often opposite) as the original example \citep{Ross2020}. Because \type{sec:counterfactuals}{Counterfactual explanations} are defined by the output class they are limited to sequence-to-class models.

Another common confusion is with \emph{counterfactual datasets}, also known as \emph{Contrast Sets}. These datasets are used in robustness research and could consist of \emph{counterfactual examples}. However, these datasets are generated without using a model \citep{Gardner2020,Kaushik2020}, and can therefore not be used to explain the model. \emph{Contrast Sets} are however important for ensuring a robust model.

In social sciences, \type{sec:counterfactuals}{counterfactual explanations} are considered highly useful for a person's ability to understand causal connections. \citet{Miller2019} explains that ``why'' questions are often answered by comparing \emph{facts} with \emph{foils}, where the term \emph{foils} is the social sciences term for \type{sec:counterfactuals}{counterfactual examples}.

\ifarxiv{\input{chapters/counterfactuals_polyjuice}}
\input{chapters/counterfactuals_mice}

\subsection{Discussion}

\paragraph{Groundedness} While \type{sec:counterfactuals}{counterfactual examples} are great for \measure{human-grounded} explanation, they struggle with \measure{functionally-groundedness}. The challenge comes from the desirables. On one side, a desirable is to provide a counterfactual example with the opposite gold label, an objective that is independent of the model. Simultaneously the search procedure should be directed by the model behavior. These objectives can at times appear opposite, although \method{sec:counterfactuals:mice}{MiCE} provide a great example of how it can be done.

\paragraph{Future work} Because the motivation for \type{sec:counterfactuals}{counterfactual examples} is often robustness, the search procedure often becomes only weakly dependent on the model such as \method{sec:counterfactuals:polyjuice}{Polyjuice} or sometimes completly independent such as \emph{Contrast Sets}.

While robustness is a perfectly valid research objective, we recommend being careful when using both robustness with interpretability to motivate the same method, as this often leads to \measure{functionally-groundedness} issues. We would therefore advocate for more counterfactual research which focuses only on interpretability and \measure{functionally-groundedness}.



%% file: chapters/counterfactuals_polyjuice.tex
\subsection{Polyjuice}
\label{sec:counterfactuals:polyjuice}

\method{sec:counterfactuals:polyjuice}{Polyjuice} by \citet{Wu2021} is primarily a \emph{counterfactual dataset} generator, and the generation is therefore detached from the model. However, by strategically filtering these generated examples such that the model's prediction is changed the most, they condition the \emph{counterfactual} generation on the model, thereby making a \posthoc{post-hoc} explanation.

The generation is done by fine-tuning a GPT-2 model \citep{Radford2019} on existing \emph{counterfactual datasets} \citep{Kaushik2020,Gardner2020,Zhang2019,Sakaguchi2019,Wieting2018,ThomasMcCoy2020}. For each pair of original and counterfactual example, they produce a training prompt, see \eqref{fig:counterfactuals:polyjuice:training-prompt} for the exact structure. What the conditoning code is and what is replaced in \eqref{fig:counterfactuals:polyjuice:training-prompt} is determined by the existing \emph{counterfactual datasets}.

\begin{equation}
\begin{aligned}
    prompt = \text{``}&\underbrace{\text{It is great for kids}}_{\text{original sentence}}
    \ \texttt{<GENERATE>} \\
    &\ \underbrace{\texttt{[negation]}}_{\text{conditioning code}}
    \ \underbrace{\text{It is \texttt{[BLANK]} great for \texttt{[BLANK]}}}_{\text{masked counterfactual}} \\
    &\ \texttt{<REPLACE>}\ 
    \underbrace{\text{not \texttt{[ANSWER]} children \texttt{[ANSWER]}}}_{\text{masking answers}}
    \ \texttt{<EOS>}\text{''}
\end{aligned}
\label{fig:counterfactuals:polyjuice:training-prompt}
\end{equation}

For \emph{counterfactual} generation, they specify the original sentence and optionally the condition code, and then let the model generate the \emph{counterfactuals}. These \emph{counterfactuals} are independent of the model. To make them dependent on the model, they filter the \emph{counterfactuals} and select those examples that change the prediction the most. One important detail is that they adjust the prediction change with an \type{sec:input-features}{importance measure} (\method{sec:input-features:shap}{SHAP}), such that the \type{sec:counterfactuals}{counterfactual examples} that could have been generated by an \type{sec:input-features}{importance measure} are valued less. An example of this explanation can be seen in \Cref{fig:counterfactuals:polyjuice}.

\begin{figure}[H]
    \centering
    \examplefigure{polyjuice}
    \caption{Hypothetical results of \method{sec:counterfactuals:polyjuice}{Polyjuice}, showing how some words were either replaced or removed to produce \type{sec:counterfactuals}{counterfactual examples}.}
    \label{fig:counterfactuals:polyjuice}
\end{figure}

To validate \method{sec:counterfactuals:polyjuice}{Polyjuice}, for a \measure{human-grounded} experiment, they show that humans were unable to predict the model's behavior for the \type{sec:counterfactuals}{counterfactual examples}, thereby concluding that their method highlights potential robustness issues. Whether \method{sec:counterfactuals:polyjuice}{Polyjuice} is \measure{functionally-grounded} is somewhat questionable, because the model is not a part of the generation process itself, it is merely used as a filtering step.

%% file: chapters/counterfactuals_mice.tex
\subsection{MiCE}
\label{sec:counterfactuals:mice}

Like \method{sec:counterfactuals:polyjuice}{Polyjuice} \citep{Wu2021}, \method{sec:counterfactuals:mice}{MiCE} \citep{Ross2020} also uses an auxiliary model to generate \emph{counterfactuals}. However, unlike \method{sec:counterfactuals:polyjuice}{Polyjuice}, \method{sec:counterfactuals:mice}{MiCE} does not depend on auxiliary datasets and the counterfactual generation is more tied to the model being explained, rather than just using the model's predictions to filter the \type{sec:counterfactuals}{counterfactual examples}.

The counterfactual generator is a T5 model \citep{Raffel2020}, a sequence-to-sequence model, which is fine-tuned by input-output-pairs, where the input consists of the gold label and the masked sentence, while the output is the masking answer, see \eqref{fig:counterfactuals:mice:training-prompt} for an example. 

\begin{equation}
\begin{aligned}
    input &= \text{``}\texttt{label:}
    \ \underbrace{\text{positive}}_{\text{gold label}}
    \texttt{,}
    \ \texttt{input:}
    \ \underbrace{\text{This movie is \texttt{[BLANK]}!}}_{\text{masked sentence}}\text{''} \\
    target &= \text{``}\texttt{[CLR]}
    \underbrace{\text{really great}}_{\text{masking answer}}
    \texttt{[EOS]}\text{''}
\end{aligned}
\label{fig:counterfactuals:mice:training-prompt}
\end{equation}

The \method{sec:counterfactuals:mice}{MiCE} approach to selecting which tokens to mask is to use an \type{sec:input-features}{importance measure}, specifically  \method{sec:input-features:gradient}{the gradient w.r.t. the input}, and then mask the top x\% most important consecutive tokens.

For generating counterfactuals, \method{sec:counterfactuals:mice}{MiCE} again masks tokens based on the \type{sec:input-features}{importance measure}, but then also inverts the gold label used for the T5-input \eqref{fig:counterfactuals:mice:training-prompt}. This way the model will attempt to infill the mask, such that the sentence will have an opposite semantic meaning. This process is then repeated via a beam-search algorithm which stops when the model prediction changes, an example of this can be seen in \Cref{fig:counterfactuals:mice}.

\begin{figure}[h]
    \centering
    \examplefigure{mice}
    \caption{Hypothetical visualization of how \method{sec:counterfactuals:mice}{MiCE} progressively creates a counterfactual $\tilde{\mathbf{x}}$ from an original sentence $\mathbf{x}$. The highlight shows the \method{sec:input-features:gradient}{gradient}  $\nabla_\mathbf{x} f(\mathbf{x};\theta)_y$, which \method{sec:counterfactuals:mice}{MiCE} uses to know what tokens to replace.}
    \label{fig:counterfactuals:mice}
\end{figure}

Because \method{sec:counterfactuals:mice}{MiCE} uses the model prediction to stop the beam-search, it will inherently be somewhat \measure{functionally-grounded}. However, it may be that using the \method{sec:input-features:gradient}{gradient} as the \type{sec:input-features}{importance measure}, is not \measure{functionally-grounded}. \citet{Ross2020} validate that using the \method{sec:input-features:gradient}{gradient} is \measure{functionally-grounded}, by looking at the number of edits and fluency of \method{sec:counterfactuals:mice}{MiCE} and compare it to a version of \method{sec:counterfactuals:mice}{MiCE} where random tokens are masked. They find that using the \method{sec:input-features:gradient}{gradient} significantly improves both fluency and reduces the number of edits it takes to change a prediction.

%% file: chapters/natural_language.tex
\section{Natural Language}
\label{sec:natural-language}

A common concern for many of the explanation methods presented in this survey is that they are difficult to understand for people without specialized knowledge. It is therefore attractive to directly generate an explanation in the form of \type{sec:natural-language}{natural language}, which can be understood by simply reading the explanation for a given example. Because these utterances explain just a single example, they are a \category{local explanation}. 

Most research in the area of \type{sec:natural-language}{natural language} explanation uses the explanations to improve the predictive performance of the model itself. The idea is that by enforcing the model to reason about its behavior, the model can generalize better \citep{Lei2016,Camburu2018,Liu2019a,Rajani2019,Kumar2020,Latcinnik2020}. These approaches are however in the category of \intrinsic{intrinsic} methods. While those methods are often quite general, they are not discussed in this survey which focuses on \posthoc{post-hoc} methods.

These \posthoc{post-hoc} methods are referred to as \emph{rationalization} methods, in the sense that they attempt to explain after a prediction has been made \citep{Rajani2019}. Note that the term is a misnomer, as rationalizations in the dictionary sense\footnote{``the action of attempting to explain or justify behaviour or an attitude with logical reasons, even if these are not appropriate.'' -- Oxford Defintion of \textit{rationalization}.} can also be false.

\input{chapters/natural_language_cage}

\subsection{Discussion}

\paragraph{Groundedness} This sub-field of natural \type{sec:natural-language}{natural language} explanations have received criticism in NLP for not evaluating \measure{functionally-grounded} \citep{Hase2020}. This issue is even more problematic because the annotated explanations are provided by humans who have no insights into the model's behavior \citep{Wiegreffe2021a}. The explanation model therefore just learns about humans' thought processes rather than the model's logical process. This issue is somewhat unique to the NLP literature and is better treated in other fields \citep{Andreas2017}.

\paragraph{Future work} Most work on natural \type{sec:natural-language}{natural language} explanations uses \intrinsic{intrinsic} methods, under the motivation that forcing the model to ``reason about itself'' will make it more accurate. Unfortunately, this hypothesis has received criticism because the little \posthoc{post-hoc} work there exist, show that this is not the case. Additionally, there are theoretical arguments for why this would not be the case \citep{Jang2021}.







%% file: chapters/natural_language_cage.tex
\subsection{Rationalizing Commonsense Auto-Generated Explanations (CAGE)}
\label{sec:natural-language:cage}

\citet{Rajani2019} provide explanations to the Common sense Question Answering (CQA) dataset, which is a multiple-choice question answering dataset \citep{Talmor2019}. The explanations are independent of the model and are provided via Amazon Mechanical Turk. To provide rationalization explanations, they then fine-tune a GPT model \citep{Alec2018}, using the question, answers, and explanation. See \eqref{fig:natural-language:cage:training-prompt} for an example of the exact prompt construction. To clarify, this GPT model is not the explained model but provides the explanations, this is known as an explainer-model.

\begin{equation}
\begin{aligned}
input &= \text{``}
\underbrace{\text{What could people do that involves talking?}}_{\text{question}}
\ \underbrace{\text{confession}}_{\text{choice 1}}
\texttt{,}
\ \underbrace{\text{carnival}}_{\text{choice 2}} \\
&\phantom{= \text{`` }}\texttt{, or}
\ \underbrace{\text{state park}}_{\text{choice 3}}
\texttt{?} 
\ \underbrace{\text{confession}}_{\text{answer}}
\ \texttt{because }
\text{''} \\
target &= \text{``}
\underbrace{\text{confession is the only vocal action.}}_{\text{rational explanation}}
\text{''}
\end{aligned}
\label{fig:natural-language:cage:training-prompt}
\end{equation}

For simpler tasks, such as ``Stanford Sentiment Treebank'' \citep{Socher2013a}, the prompt could simply be ``\texttt{[input]. [answer] because [explanation]}'', see \Cref{fig:natural-language:cage} for hypothetical explanations using such a setup. Because \method{sec:natural-language:cage}{CAGE} uses a generative model, where \texttt{[answer]} can be a sequence, it is not limited to sequence-to-class problems.
 
\begin{figure}[h]
    \centering
    \examplefigure{cage}
    \caption{Hypothetical explanations from using \method{sec:natural-language:cage}{CAGE} to produce rationalizations for the prediction.}
    \label{fig:natural-language:cage}
\end{figure}

They find that rationalization explanations provide nearly identical explanations as reasoning explanations (those where the answer is not known by the explanation model). The method is validated to be \measure{human-grounded}, by tasking humans to use the explanation to predict the model behavior, again they find identical performance.

It is questionable if \method{sec:natural-language:cage}{CAGE} is  \measure{functionally-grounded}, as its only connection to the explained model is during inference, where the \texttt{answer} is produced by the explained model. Because there are no other connections to the explained model, their is little reason to think the GPT explainer-model can reflect the models behavior. If the humans who provided explanations had specialist insight into the model, then an argument could be made for \method{sec:natural-language:cage}{CAGE} to be  \measure{functionally-grounded}. However, as the humans were Mechanical Turk workers, this is unlikely.

%% file: chapters/concepts.tex
\section{Concepts}
\label{sec:concepts}

A \type{sec:concepts}{concept explanation} attempts to explain the model, in terms of an abstraction of the input, called a \type{sec:concepts}{concept}. A classical example in computer vision, is to explain how the concept of stripes affects the classification of a zebra. Understanding this relationship is important, as a computer vision model could classify a zebra based on a horse-like shape and a savana background. Such relation may yield a high accuracy score but is logically wrong.

The term \type{sec:concepts}{concept} is much more common in computer vision \citep{Goyal2019,Kim2018,Mu2020} than in NLP. Instead, the subject is often framed more concretly as bias-detection, in NLP. For example, \citet{Vig2020a} uses the concept of occupation-words like \emph{nurse}, and relates it to the classification of the words \emph{he} and \emph{she}.

Regardless of the field, in both NLP and CV, only a single class or small subset of classes are analyzed. For this reason, \type{sec:concepts}{concept explanation} belong in its own category of \category{class explanations}. However, in the future, we will likely see more types of \category{class explanations}.

\input{chapters/concepts_nie}

\subsection{Discussion}

\paragraph{Groundedness} As a new field, there is not much work on \measure{groundedness}. \citet{Vig2020a} do not measure either \measure{functionally-groundedness} or \measure{human-groundedness} on \method{sec:concepts:natural-indirect-effect}{Natural Indirect Effect}. It is also not obvious how \measure{functionally-groundedness} could be measured. Note, that this situation is not unique to \type{sec:concepts}{concept explanation}, as many other communication appraoches also don't have an established measure of \measure{functionally-groundedness}.

\paragraph{Future work} \type{sec:concepts}{Concept explanation} requires either a new dataset or annotation of an existing dataset. This can be quite expensive and impractical, especially when there is no concrete concept in mind and the user wants a more exploratory explanation. However, there is new research towards discovering concepts automatically \citep{Ghorbani2019c}.

%% file: chapters/concepts_nie.tex
\subsection{Natural Indirect Effect (NIE)}
\label{sec:concepts:natural-indirect-effect}

Consider a language model with the prompt $\mathbf{x} = \text{``The nurse said that''}$. To measure if the gender-stereotype of ``nurse'' is female, it is natural to compare $p(\text{she} | \mathbf{x};\theta)$ with $p(\text{he} | \mathbf{x};\theta)$, or alternatively $p(\text{they} | \mathbf{x};\theta)$. Generalized, \citet{Vig2020a} express this as 
\begin{equation}
    \operatorname{bias-effect}(\mathbf{x};\theta) = \frac{p(\text{\emph{anti-stereotypical}} | \mathbf{x};\theta)}{p(\text{\emph{stereotypical}} | \mathbf{x};\theta)} \ .
\end{equation}

\citet{Vig2020a} then provide insight into which parts of the model are responsible for the bias. They do this by measuring the \method{sec:concepts:natural-indirect-effect}{Natural Indirect Effect} (NIE) from causal mediation analysis. Although this appraoch applies to a sequence-to-sequence model, only one token being considered at a time. It is therefore possible also apply it to purely sequence-to-class models.

Given a model $f(\mathbf{x};\theta)$, mediation analysis is used to understand how a latent representation $z(\mathbf{x};\theta)$ (called the mediator) affects the final model output. This latent representation can either be a single neuron or several neurons, like an attention head. The \method{sec:concepts:natural-indirect-effect}{Natural Indirect Effect} measures the effect that goes though this mediator.

To measure causality, an \emph{intervention} on the concept measured must be made. As intervention, \citet{Vig2020a} replace ``nurse'' with ``man'', or ``woman'' for oppositely biased occupations. They call this replace operation \texttt{set-gender}.

Then to measure the effect of the mediator \citet{Vig2020a} introduce
\begin{equation}
    \operatorname{mediation-effect}_{m_1,z,m_2}(\mathbf{x};\theta) = \frac{\operatorname{bias-effect}_{z(m_2(\mathbf{x});\theta)}(m_1(\mathbf{x});\theta)}{\operatorname{bias-effect}(\mathbf{x};\theta)},
\end{equation}
where $m \in \{\texttt{identity}, \texttt{set-gender}\}$ and $\operatorname{bias-effect}_{z(m_2(\mathbf{x}))}(\cdot)$ is $\operatorname{bias-effect}(\cdot)$ but uses a modified model with the mediator values for $z(m_1(\mathbf{x}))$ fixed. With this definition, the \method{sec:concepts:natural-indirect-effect}{Natural Indirect Effect} follows from causal mediation analysis literature \citep{Pearl2001}.

\begin{equation}
\begin{aligned}
    \mathrm{NIE}_z = \mathbb{E}_\mathbf{x \in \mathcal{D}}[ &\operatorname{mediation-effect}_{\texttt{identity}, z, \texttt{set-gender}}(\mathbf{x};\theta) \\
    &- \operatorname{mediation-effect}_{\texttt{identity}, z, \texttt{identity}}(\mathbf{x};\theta)]
\end{aligned}
\end{equation}

\citet{Vig2020a} apply \method{sec:concepts:natural-indirect-effect}{Natural Indirect Effect} to a small GPT-2 model, where the mediator is an attention head. By doing this, \citet{Vig2020a} can identify which attention heads are most responsible for the gender bias, when considering the occupation concept. Hypothetical results, but results similar to those presented in \citet{Vig2020a}, are presented in \Cref{fig:concepts:natural-indirect-effect}.

\begin{figure}[h]
    \centering
    \examplefigure{naturaleffect}
    \caption{Visualization of hypothetical \method{sec:concepts:natural-indirect-effect}{Natural Indirect Effect} (NIE) results, similar to \citet{Vig2020a}. Such visualization can reveal which attention-head are responsible for gender bias, in a small GPT-2 model. A stronger color indicates a higher NIE, meaning more responsible for the bias.}
    \label{fig:concepts:natural-indirect-effect}
\end{figure}

%% file: chapters/vocabulary.tex
\section{Vocabulary}
\label{sec:vocabulary}

For this category, we define the term \type{sec:vocabulary}{vocabulary explanation} as methods which explain the whole model in relation to each word in the vocabulary and is therefore a \category{global explanation}. 

In the sentiment classification context, a useful insight could be if positive and negative words are clustered together respectively. Furthermore, perhaps there are words in those clusters which can not be considered of either positive or negative sentiment. Such a finding could indicate a bias in the dataset.

Because \type{sec:vocabulary}{vocabulary explanations} explain using the model's vocabulary, they can often be applied to both sequence-to-class and sequence-to-sequence models. This is esspecially true for explainations based on the embedding matrix, which so is almost exclusively the case.

Because an embedding matrix is often used and because neural NLP models often use pre-trained word embeddings, most research on \type{sec:vocabulary}{vocabulary explanations} is applied to the pre-trained word embeddings \citep{Mikolov2013a, Pennington2014}. However, in general, these explanation methods can also be applied to the word embeddings after training.

\input{chapters/vocabulary_projection}
\input{chapters/vocabulary_rotation}

\subsection{Discussion}

\paragraph{Groundedness} In terms of \measure{human-grounded}, \type{sec:vocabulary}{vocabulary explanation} are one of the few sub-fields that have a well established test, namely the \emph{word intrusion} test \citep{Chang2009}. It is therefore hard to justify when methods in this category replace humans with an algorithm, as this largely invalidates the test.

\paragraph{Future work} While past work, such as Latent Dirichlet Allocation (LDA) \citep{Blei2003a}, have provided great \type{sec:vocabulary}{vocabulary explanations}, contemporary work using neural networks is quite limited and is mostly based on the embedding matrix. This is a pity, as the embedding matrix only provides a limited picture and it is not hard to imagine using other information sources to create \type{sec:vocabulary}{vocabulary explanations}. For example, one could aggregate the word-contributions provided by \type{sec:input-features}{input feature} explanations.

%% file: chapters/vocabulary_projection.tex
\subsection{Projection}
\label{sec:vocabulary:projection}

A common visual explanation is to project embeddings to two or three dimensions. This is particularly attractive, as word embeddings are of a fixed number of dimensions, and can therefore draw from the very rich literature on projection visualizations of tabular data, most notable is perhaps Principal Component Analysis (PCA) \citep{Pearson1901}.

\paragraph{t-SNE} Another popular and more recent method is t-SNE \citep{VanDerMaaten2008}, which has been applied to word embeddings \citep{Li2016}. This method has in particular been attractive as it allows for non-linear transformations, while still keeping points that are close in the word embedding space, also close in the visualization space. t-SNE does this by representing the two spaces with two distance-distributions, it then minimizes the KL-divergence by moving the points in the visualization space.

Note that \citet{Li2016} does not go further to validate t-SNE in the context of word embeddings, except to highlight that words of similar semantic meaning are close together, we provide a similar example in \Cref{fig:vocabulary:projection}.

\begin{figure}[h]
    \centering
    \examplefigure{projection}
    \caption{PCA \citep{Pearson1901} and t-SNE \citep{VanDerMaaten2008} projection of GloVe \citep{Pennington2014} embeddings for the words in the semantic classification examples, as shown in \Cref{sec:motivating-example} and elsewhere in this survey.}
    \label{fig:vocabulary:projection}
\end{figure}

\paragraph{Supervised projection} A problem with using PCA and t-SNE, is that they are unsupervised. Hence, while they might find a projection that offers high contrast, this projection might not correlate with what is of interest. An attractive alternative is therefore to define the projection, such that it reveals the subject of interest.

\citet{Bolukbasi2016} are interested in how gender-biased a word is. They explore gender-bias, by projecting each word onto a gender-specific vector and a gender-neutral vector. Such vectors can either be defined as the directional vector between ``he'' and ``she'', or alternative. \citet{Bolukbasi2016} also use multiple gender-specific pairs such as ``daughter-son`` and ``herself-himself'', and then use their first Principal Component as a common projection vector.

%% file: chapters/vocabulary_rotation.tex
\subsection{Rotation}
\label{sec:vocabulary:rotation}

The category of, for example, all positive sentiment words may have similar word embeddings. However, it is unlikely that a particular basis dimension describes positive sentiment itself. A useful interpretability method, is therefore to rotate the embedding space such that the basis-dimensions in the new rotated embedding space represents significant concepts. This is distinct from \method{sec:vocabulary:projection}{projection} methods because there is no loss of information as only a rotation is applied.

\citet{Park2017} perform such rotation using \emph{Exploratory Factor Analysis (EFA)} \citep{Costello2005}. The idea is to formalize a class of rotation matrices, called the \emph{Crawford-Ferguson Rotation Family} \citep{Crawford1970}. The parameters of this rotation formulation are then optimized, to make the rotated embedding matrix only have a few large values in each row or column. As an hypothetical example see \Cref{tab:vocabulary:rotation:basis-dimensions}.

\begin{table}[h]
\centering
\begin{tabular}{ll}
\toprule
Basis-dimension & top-3 words \\
\midrule
1 & handsome, feel, unpredictable \\
2 & most, best, anything \\
3 & suspense, drama, comedy \\
\bottomrule
\end{tabular}
\caption{Fictive example of the top-3 words for each basis-dimension in the rotated word embeddings.}
\label{tab:vocabulary:rotation:basis-dimensions}
\end{table}

\citet{Park2017} validate this method to be \measure{human-grounded} by using the \emph{word intrusion} test. The classical word Intrusion test \citep{Chang2009} provides 6 words to a human annotator, 5 of which should be semantically related, the 6th is the intruder which is semantically different. The human annotator then has to identify the intruder word. Importantly, semantic relatedness is in this case defined as the top-5 words of a given basis-dimension in the rotated embedding matrix.

Unfortunately, rather than having humans detect the intruder, \citet{Park2017} use a distance ratio, related to the cosine-distance, as the detector. This is problematic, as distance is directly related to how the semantically related words were chosen. In this case the intruder should have been identified either by a human or an oracle model.

%% file: chapters/ensemble.tex
\section{Ensemble}
\label{sec:ensemble}

\type{sec:ensemble}{Ensemble} explanations attempts to provide a \category{global explanation} by collecting multiple \category{local explanations}. This is done such that each \category{local explaination} represents the different modes of the model.

The extreme of this idea would be to provide a \category{local explaination} for every possible input, thereby providing a \category{global explanation}. Unfortunately, such an explanation is too much information for a human to understand and would not be \measure{human-grounded}. As \citet{Miller2019} state, an explanation should be selective. The task of \type{sec:ensemble}{ensemble} explanations, is therefore to strategically select representative examples and their corresponding \category{local explainations}.

The assumption is that the model operates within different modes. Futhermore, that one example, or a few examples, from each mode can sufficiently represent the models entire behavior. For example, in sentimate classification of movie reviews, a model may have one behavior for comments about the acting, another behavior for comments about the music score, etc.

\type{sec:ensemble}{Ensemble} explanations is a very broad category of explanations, as for every type of \category{local explanation} method there is, an \type{sec:ensemble}{ensemble} explanation could in principle be constructed. As such, if it can be applied to sequence-to-class or sequence-to-sequence models depends depends on the specific method. However, in practice very few \type{sec:ensemble}{ensemble} methods have been proposed, and most of them apply only to tabular data \citep{Ibrahim2019,Ramamurthy2020,Sangroya2020}.

\input{chapters/ensamble_splime}

\subsection{Discussion}

\paragraph{Groundedness} The \measure{functionally-groundedness} of \type{sec:ensemble}{ensemble} explanations is very much dependent on the \measure{functionally-groundedness} of the \category{local explanation}. It is therefore diffcult to imagine a general evaluation appraoch for \type{sec:ensemble}{ensemble} explanations. However, even for \category{local explanations} with established validation \measure{functionally-groundedness} does not come for free, as also the selection algorithm also needs to be validated.

\paragraph{Future work} As mentioned there is not much work using \type{sec:ensemble}{ensemble} explanations. This is because when non-tabular data is used, it is more challenging to compare the selected explanations to ensure they represent different modes. Even \method{sec:ensemble:sp-lime}{SP-LIME} \citep{Ribeiro2016} which does apply to NLP tasks, uses a Bag-of-Word representation as a tabular proxy. Additionally, we can imagine that \type{sec:ensemble}{ensemble} explanations are hard to scale, as datasets increases and models get more complex with more modes.

That being said, we would be curious to see more work in this category. For example, an \type{sec:ensemble}{ensemble} explanation which used a \type{sec:influential-examples}{influential example} method to show the overall most relevant observations.

%% file: chapters/ensamble_splime.tex
\subsection{Submodular Pick LIME (SP-LIME)}
\label{sec:ensemble:sp-lime}

\method{sec:ensemble:sp-lime}{SP-LIME} by \citet{Ribeiro2016} attempts to select $B$ observations (a budget), such that they represent the most important features based on their \method{sec:input-features:lime}{LIME} explanation. Note that, while \method{sec:input-features:lime}{LIME} explanations can be made for each output token and can therefore be used in a sequence-to-sequence context, \method{sec:ensemble:sp-lime}{SP-LIME} do assume a sequence-to-class model.

\method{sec:ensemble:sp-lime}{SP-LIME} calculates the importance of each feature $v$, by summing the absolute importance for all observations in the dataset, this total importance is $\mathbf{I}_v$ in \eqref{eq:ensemble:sp-lime:def}. The objective is then to maximize the sum of $\mathbf{I}_v$ given a subset of features, by strategically selecting $B$ observations. Note that selecting multiple observations which represent the same features will not improve the objective. The specific objective is formalized in \eqref{eq:ensemble:sp-lime:def}, which \citet{Ribeiro2016} optimize greedily.

\begin{equation}
\begin{aligned}
\mathbf{G}_{\text{SP-LIME}} = &\argmax_{\tilde{\mathcal{D}} \text{ s.t. } |\tilde{\mathcal{D}}| \le B} \sum_{v=1}^{V} \mathds{1}_{\left[\exists \tilde{\mathbf{x}}_i \in \tilde{\mathcal{D}}\,:\, \left|\mathbf{E}_{\operatorname{LIME}}\left(\tilde{\mathbf{x}}_i, \argmax_i p(i|\tilde{\mathbf{x}}_i;\theta)\right)_v\right| > 0\right]} \mathbf{I}_v \\
&\text{where } \tilde{\mathcal{D}} \subseteq \mathcal{D} \\
&\phantom{\text{where }} \mathbf{I}_v = \sum_{\tilde{\mathbf{x}}_i \in \mathcal{D}} \left|\mathbf{E}_{\operatorname{LIME}}\left(\tilde{\mathbf{x}}_i, \argmax_i p(i|\tilde{\mathbf{x}}_i;\theta)\right)_v\right|
\end{aligned}
\label{eq:ensemble:sp-lime:def}
\end{equation}

\begin{figure}[h]
    \centering
    \examplefigure{splime}
    \caption{Visualization of \method{sec:ensemble:sp-lime}{SP-LIME} in a hypothetical setting. The matrix shows how each selected observation represents the different modes of the model. The left-side shows two out of the four selected example and their \method{sec:input-features:lime}{LIME} explanation.}
    \label{fig:ensemble:sp-lime}
\end{figure}

A major challenge with \method{sec:ensemble:sp-lime}{SP-LIME} is that it requires computing a \method{sec:input-features:lime}{LIME} explanation for every observation. Because each \method{sec:input-features:lime}{LIME} explanation involves optimizing a logistic regression this can be quite expensive. To reduce the number of observations that need to be explained, \citet{Sangroya2020} proposed using \emph{Formal Concept Analysis} to strategically select which observations to explain. However, this approach has not yet been applied to NLP.

\citet{Ribeiro2016} validate \method{sec:ensemble:sp-lime}{SP-LIME} to be \measure{human-grounded} by asking humans to select the best classifier, where a ``wrong classifier'' is trained on a biased dataset and a ``correct classifier'' is trained on a curated dataset. \citet{Ribeiro2016} then compare \method{sec:ensemble:sp-lime}{SP-LIME} with a random baseline, which simply selects random observations. From this experiment, they find that 89\% of humans can select the best classifier using \method{sec:ensemble:sp-lime}{SP-LIME}, where as only 75\% can select the best classifier based on the random baseline.

%% file: chapters/linguistic_information.tex
\section{Linguistic Information}
\label{sec:linguistic-information}

To validate that a natural language model does something reasonable, a popular approach is to attempt to align the model with the large body of linguistic theory that has been developed for hundreds of years. Because these methods summarize the model, they are a case of \category{global explanation}.

Methods in this category either probe by strategically modifying the input to observe the model's reaction or show alignment between a latent representation and some linguistic representation. The former is called \method{sec:linguistic-information:behavioral-probes}{behavioral probes} or \method{sec:linguistic-information:behavioral-probes}{behavioral analysis}, the latter is called \method{sec:linguistic-information:structural-probes}{structural probes} or \method{sec:linguistic-information:structural-probes}{structural analysis}. Which type of models these strategies applies to depends on the specific method. However, in general \method{sec:linguistic-information:behavioral-probes}{behavioral probes} applies primarily to sequence-to-class models and \method{sec:linguistic-information:structural-probes}{structural probes} applies to both sequence-to-class and sequence-to-sequence models.

One especially noteworthy subcategory of \method{sec:linguistic-information:structural-probes}{Structural Probes} is \emph{BERTology}, which specifically focuses on explaining the BERT-like models \citep{Devlin2019,Liu2019,Brown2020}. BERT's popularity and effectiveness have resulted in countless papers in this category \citep{Michel2019,Coenen2019,Clark2019a,Rogers2020,Tenney2019a}, hence the name \emph{BERTology}. Some of the works use the attention of BERT and are therefore \intrinsic{intrinsic} explanations, while others simply probe the intermediate representations and are therefore \posthoc{post-hoc} explanations.

There already exist well-written survey papers on \type{sec:linguistic-information}{Linguistic Information} explanations. In particular, \citet{Belinkov2020} cover \method{sec:linguistic-information:behavioral-probes}{behavioral probes} and \method{sec:linguistic-information:structural-probes}{structural probes}, \citet{Rogers2020} discuss \emph{BERTology}, and \citet{Belinkov2019} cover \method{sec:linguistic-information:structural-probes}{structural probing} in detail. In this section, we will therefore not go in-depth, but simply provide enough context to understand the field and importantly mention some of the criticisms, that we believe have not been sufficiently highlighted by other surveys.

\input{chapters/linguistic_information_behavioral}
\input{chapters/linguistic_information_structual}

\subsection{Discussion}

\paragraph{Groundedness} Considering the vast amount of research on \type{sec:linguistic-information}{linguistic information} explanations, we find it worrying that there isn't more work on evaluating if these explanations are actually useful, in terms of the \measure{human-groundedness} and \measure{functionally-groundedness}. Without such evaluation, it is difficult to ensure that the field of \type{sec:linguistic-information}{linguistic information} explanations moves in a productive direction.

\paragraph{Future work} Considering the \measure{groundedness} issues in \type{sec:linguistic-information}{linguistic information} explanations, we advocate for more focus on \measure{groundedness}. \citet{Voita2020} provide a great solution to how the \measure{functionally-groundedness} issues can be overcome. However, the field still lacks independent study on \measure{human-groundedness} and \measure{functionally-groundedness}.

%
%






%% file: chapters/linguistic_information_behavioral.tex
\subsection{Behavioral Probes}
\label{sec:linguistic-information:behavioral-probes}

The research being done in \method{sec:linguistic-information:behavioral-probes}{behavioral probes}, also called \method{sec:linguistic-information:behavioral-probes}{behavioral analysis}, is not just for interpretability but also to measure the robustness and generalization ability of the model. For this reason, many \emph{challenge datasets} are in the category of \method{sec:linguistic-information:behavioral-probes}{behavioral analysis}. These datasets are meant to test the model's generalization capabilities, often by containing many observations of underrepresented modes in the training datasets. However, the model's performance on \emph{challenge datasets} does not necessarily provide interpretability.

One of the initial papers providing interpretability via \method{sec:linguistic-information:behavioral-probes}{behavioral probes} is that by \citet{Linzen2016}. They probe a language model's ability to reason about subject-verb agreement correctly. A recent work, by \citet{Sinha2021, Clouatre2021a}, find that destroying syntax by shuffling words does not significantly affect a model trained on an NLI task, indicating that the model does not achieve natural language understanding.

As mentioned, this area of research is quite large and \citet{Belinkov2020} cover \method{sec:linguistic-information:behavioral-probes}{behavioral probes} in detail. Therefore, we just briefly discuss the work by \citet{ThomasMcCoy2020}, which provide a particularly useful example on how \method{sec:linguistic-information:behavioral-probes}{behavioral probes} can be used to provide interpretability.

\citet{ThomasMcCoy2020} look at Natural Language Inference (NLI), a task where a premise (for example, ``The judge was paid by the actor'') and a hypothesis (for example, ``The actor paid the judge'') are provided, and the model should inform if these sentences are in agreement (called \emph{entailment}). The other options are \emph{contradiction} and \emph{neutral}. \citet{ThomasMcCoy2020} hypothesise that models may not actually learn to understand the sentences but merely use heuristics to identify \emph{entailment}.

They propose 3 heuristics based on the linguistic properties: lexical overlap, subsequence, and constituent. An example of lexical overlap is the premise ``\textbf{The} \textbf{doctor} was \textbf{paid} by \textbf{the} \textbf{actor}'' and hypothesis ``The doctor paid the actor''. The proposed heuristic is that this observation would be classified as \emph{entailment} by the model due to lexical overlap, even though this is not the correct classification.

To test for these heuristics, \citet{ThomasMcCoy2020} developed a dataset, called HANS, which contains examples with these linguistic properties but do not have \emph{entailment}. The results (\cref{tab:linguistic-information:behavioral-probes:performance}) validates the hypothesis that the model relies on these heuristics rather than a true understanding of the content. Had just an average score across all heuristics been provided, this would just be a robustness measure. However, by providing meta-information on which pattern each observation follows, the accuracy scores provide interpretability on where the model fails.
\begin{table}[h]
    \centering
    \begin{tabular}{ccccc}
        \toprule
        & Lexical Overlap & Subsequence & Constituent & Average \\
        \midrule
        BERT \citep{Devlin2019} & 17\% & 5\% & 17\% & -- \\
        Human (Mechanical Turk) & -- & -- & -- & 77\% \\
        \bottomrule
    \end{tabular}
    \caption{Performance on the HANS dataset provided by \citet{ThomasMcCoy2020}. Unfortunately, \citet{ThomasMcCoy2020} do not provide enough information to make a direct comparision possible. For comparison, BERT has 83\% accuracy on MNLI \citep{Williams2018}, which was used for training.}
    \label{tab:linguistic-information:behavioral-probes:performance}
\end{table}

In terms of \emph{functionally-groundedness}, \citet{ThomasMcCoy2020} perform no explicit evaluation. However, given that \method{sec:linguistic-information:behavioral-probes}{behavioral probes} merely evaluate the model, \emph{functionally-groundedness} is generally not a concern. Furthermore, while \citet{ThomasMcCoy2020} do evaluate with humans, this is not a \emph{human-grounded} evaluation. Because they only use humans to evaluate the dataset, not if the explanation itself is suitable to humans.

%% file: chapters/linguistic_information_structual.tex
\subsection{Structural Probes}
\label{sec:linguistic-information:structural-probes}

Probing methods primarily use a simple neural network, often just a logistic regression, to learn a mapping from an intermediate representation to a linguistic representation, such as the Part-Of-Speech (POS).

One of the early papers, by \citet{Shi2016}, analyzed the sentence-embeddings of a sequence-to-sequence LSTM, by looking at POS (part-of-speech), TSS (top-level syntactic sequence), SPC (the smallest phrase constituent for each word), tense (past or non-past), and voice (active or passive). Similarly, \citet{Adi2017} used a multi-layer-perceptron (MLP) to analyze sentence-embeddings for sentence-length, word-presence, and word-order. More recently \citet{Conneau2018} have been using similar linguistic tasks and MLP probes but have extended previous analyses to multiple models and training methods.

Analog to these papers, a few methods use cluster algorithms instead of logistic regression \citep{Brunner2019}. Additionally, some methods only look at \emph{word embeddings} \citep{Kohn2015}. The list of papers is very long, we suggest looking at the survey paper by \citet{Belinkov2019}.

\paragraph{BERTology} As an instructive example of probing in BERTology, the paper by \citet{Tenney2019a} is briefly described. Note that this is just one example of a vast number of papers. \citet{Rogers2020} offer a much more comprehensive survey on BERTology.

\citet{Tenney2019a} probe a BERT model \citep{Devlin2019} by computing a learned weighted-sum $\mathbf{z}_i(\mathbf{x};\theta)$ for each intermediate representation $\mathbf{h}_{l,i}(\mathbf{x};\theta)$ of the token $i$, as described in \eqref{eq:linguistic-information:probing:bert-sum}.
\begin{equation}
\begin{aligned}
    \mathbf{z}_i(\mathbf{x};\theta) &= \gamma \sum_{l=1}^L s_l \mathbf{h}_{l,i}(\mathbf{x};\theta) \\
    \text{where } \mathbf{s} &= \operatorname{softmax}(\mathbf{w})
\end{aligned}
\label{eq:linguistic-information:probing:bert-sum}
\end{equation}

The weighted-sum $\mathbf{z}_i(\mathbf{x})$ is then used by a classifier \citep{Tenney2019}, and the weights $s_l$, parameterized by $\mathbf{w}$, describe how important each layer $l$ is. The results can be seen in \Cref{fig:linguistic-information:probing:bertology}.

\begin{figure}[h]
    \centering
    \examplefigure{probes}
    \caption{Results by \citet{Tenney2019a} which shows how much each BERT  \citep{Devlin2019} layer is used for each linguistic task. The $F_1$ score for each task is also presented.}
    \label{fig:linguistic-information:probing:bertology}
\end{figure}

\paragraph{Criticisms} A growing concern in the field of probing methods is that given a sufficiently high-dimensional embedding, complex probe, and large auxiliary dataset, the probe can learn everything from anything. If this concern is valid, it would mean that the probing methods do not provide \measure{functionally-grounded explanations} \citep{Belinkov2021}.

Recent work attempts to overcome this concern by developing baselines. \citet{Zhang2018} suggest learning a probe from an untrained model, as a baseline. In that paper, they find probes can indeed achieve high accuracy from an untrained model unless the auxiliary dataset size is decreased dramatically. Similarly, \citet{Hewitt2019} use randomized datasets as a baseline, called a control task. For example, for POS they assign a random POS-tag to each word, following the same empirical distribution of the non-randomized dataset. They find that equally high accuracy can be achieved on the randomized dataset unless the probe is made extraordinarily small.

\paragraph{Information-Theoretic Probing} The solutions presented by \citet{Zhang2018} and \citet{Hewitt2019} are useful. However, limiting the probe and dataset size could make it impossible to find complex hidden structures in the embeddings.

\citet{Voita2020} attempt to overcome the criticism by a more principled approach, using information theory. More specifically, they measure the required complexity of the probe as a communication effort, called \emph{Minimum Description Length} (MDL), and compare the MDL with a control task similar to \citet{Hewitt2019}. They find, similar to \citet{Hewitt2019}, that the probes achieve similar accuracy on the probe dataset as on the control task. However, the control task is much harder to communicate (the MDL is higher), indicating that the probe is much more complex, compared to training on the probe dataset.

%% file: chapters/rules.tex
\section{Rules}
\label{sec:rules}

\type{sec:rules}{Rule} explanations attempt to explain the model by a simple set of rules, therefore they are an example of \category{global explanations}.

Reducing highly complex models like neural networks to a simple set of rules is likely impossible. Therefore, methods that attempt this simplify the objectivity by only explaining one particular aspect of the model.

Due to the challenges of producing rules, there is little research attempting it. We will present \method{sec:rules:comp-explain-neuron}{Compositional Explanations of Neurons} \citep{Mu2020} and \method{sec:rules:sear}{SEAR} \citep{Ribeiro2018}.

\input{chapters/rules_sear}
\ifarxiv{\input{chapters/rules_compositional}}

\subsection{Discussion}

\paragraph{Future work} As mentioned, there is little work on \type{sec:rules}{rule} explanations. While this is definetly due to the inherent challenge, it is not too hard to imagine something like the \method{sec:input-features:anchors}{Anchor} method be modified towards \category{global explanation}, in which case it would be a \type{sec:rules}{rule} explanation.

\paragraph{Groundedness} Because the category of \type{sec:rules}{rule} explanations can be very diverse, \measure{groundedness} evaluation would likely depend on the specific explanation method. However, generally \measure{functionally-groundedness} can be measured by asserting if the rule holds true by evaluating it on the dataset and compare with the model response. Additionally, \measure{human-groundedness} can be evaluated by asking humans to predict the model's output or choose the better model.

%% file: chapters/rules_sear.tex
\subsection{Semantically Equivalent Adversaries Rules (SEAR)}
\label{sec:rules:sear}

\method{sec:rules:sear}{SEAR} is an extension of the \method{sec:adversarial-examples:sea}{Semantically Equivalent Adversaries} (SEA) method \citep{Ribeiro2018}, where they developed a sampling algorithm for finding adversarial examples. Hence, the rule-generation objective is simplified, as only rules that describe what breaks the model needs to be generated. Additionally, because \method{sec:rules:sear}{SEAR} uses an \type{sec:adversarial-examples}{adversarial examples} explanation, it only applies to sequence-to-class models.

\begin{figure}[h]
    \centering
    \examplefigure{sear}
    \caption{Hypothetical example showing rules which commonly break the model. The flip-rate describes how often these rules break the model. $\mathbf{x}$ represents the original sentence and $\tilde{\mathbf{x}}$ represents an adversarial example.}
    \label{fig:rules:sear}
\end{figure}

\citet{Ribeiro2018} propose rules by simply observing individual word changes found by the \method{sec:adversarial-examples:sea}{SEA} method discussed earlier, and then compute statistics on the bi-grams of the changed word and the Part of Speech of the adjacent word, \Cref{fig:rules:sear} shows examples of this. If the proposed rule has a high success-rate (called filp-rate), in terms of providing a semantically equivalent adversarial sample, it is considered a rule.

The authors validate this approach by asking experts to produce rules, and then compare the success-rate of human-generated rules and \method{sec:rules:sear}{SEAR}-generated rules. They find that the rules generated by \method{sec:rules:sear}{SEAR} have a higher success-rate.

%% file: chapters/rules_compositional.tex
\subsection{Compositional Explanations of Neurons}
\label{sec:rules:comp-explain-neuron}

In \method{sec:rules:comp-explain-neuron}{Compositional Explanations of Neurons} by \citet{Mu2020}, the rule generation problem is simplified by only relating the presence of input words to the activation of a single neuron.

The rules typically have the form of logical rules, meaning \texttt{not}, \texttt{and}, and \texttt{or}, where the booleans indicate a word is present, although \citet{Mu2020} do not make any hard constraints here. For example, in an NLI task they also have indicators for POS-presence and word-overlap between the hypothesis and premise. If these rules are satisfied it means the neuron activation is above a defined threshold. For example, in a $\operatorname{ReLU}(\cdot)$ unit one can threshold if its post-activation is above 0.

\begin{figure}[h]
    \centering
    \examplefigure{comp}
    \caption{Hypothetical example showing rules which activates a selected neuron. $\operatorname{IoU}$ is how often the rule activated the neuron, compared to cases where either the rule is true or the neuron activated (higher is better).}
    \label{fig:rules:comp}
\end{figure}

Given a dataset $\mathcal{D}$, a neuron activation $z_n(\mathbf{x})$, a threshold $\tau$, and a indicator function for the rule $R(\mathbf{x})$, the the aggrement between the rule and the neuron activation can be measured with the \emph{Intersection over Union score}:
\begin{equation}
    \operatorname{IoU}(n, R) = \frac{\sum_{x \in \mathcal{D}} \mathds{1}(z_n(\mathbf{x}) > \tau \land R(\mathbf{x}))}{\sum_{x \in \mathcal{D}} \mathds{1}(z_n(\mathbf{x}) > \tau \lor R(\mathbf{x}))}
\end{equation}

For one particular neuron $n$, the combinatorial rule $R$ is then constructed using beam-search which stops at a pre-defined number of iterations. At each iteration, all feature indicator functions (e.g. word in $\mathbf{x}$) and their negative, combined with the logical operators \texttt{and} and \texttt{or}, are scored using $\operatorname{IoU}(n, R)$.

Unfortunately, \citet{Mu2020} do not perform any \measure{groundedness} validation of this approach. Furthermore, as the method only looks at the relation between the input and the neuron, it is unclear how much the selected neuron affects the output.

%% file: chapters/limitations.tex
\section{Limitations}
\label{sec:limitations}

While it is the goal of this survey to provide an overview and categorization of current post-hoc interpretability for neural NLP models, we also recognize that the field is too vast to include all works in this survey. To decide what works to include, the overall has been to focus on diversity in terms of communication approach and information used. Essentially, to make \Cref{tab:overview} as comprehensive as possible.

Communication approaches like \type{sec:input-features}{input features} and \type{sec:linguistic-information}{lingustic information} have a particularly large amount of literature, which we did not discuss, as that would outweigh other communication approaches. For these two approaches we focus on highlighting the progression of the field.

Beyond this overarching limitation, the following two limitations are worth discussing.

\paragraph{Quantitative comparisons} Ideally, this survey would include quantitative comparisons of the methods. However, there currently does not exist an unified and principled benchmark yet. Producing a principal benchmark is in itself extreamly difficult and out of scope for this survey, in \Cref{sec:future-directions} we discuss further where this difficulty comes from. Performing quantitative comparisons would therefore best be left for future work on interpretability benchmarks.

\paragraph{Visual examples} Because communication is essential to this survey, visual examples of how the method communicates have been provided throughout this survey. These examples are however fictive and optimistic, showing often the best case for each explanation method. However, in practice, accurate and highly useful explanations can only be produced for some examples for \category{local explanations}, or some datasets in the case of \category{class and global explanations}. Furthermore, the visualizations are not necessarily the most effective visualizations but are instead what we believe to be the most canonical visualizations.

How an explanation method should be visualized is its own field of study and should draw from human-computer interface literature. This is something that was not covered in this survey.

%% file: chapters/findings.tex
\section{Findings}
\label{sec:findings}

This survey covers a large range of methods. In particular, we discuss how each method communicates and is evaluated. However, some discussion is not specific to any motivation, measure, or method for interpretability. Therefore, this section covers a few valuable findings which should be discussed from a holistic perspective.

\paragraph{Terminology} Because interpretability is an emerging field, terminology still varies significantly from paper to paper. In particular, the terminology regarding measures of interpretability vary. For example, \measure{human-groundedness} is often confused with \measure{functionally-groundedness}, and for each measure category there are synonyms such as \measure{simulatability}, and \measure{comprehensibility} for \measure{human-groundedness}. Additionally, the terms for the communication types are sometimes confused. Especially, \type{sec:adversarial-examples}{adversarial examples} and \type{sec:counterfactuals}{counterfactuals} are occasionally interchanged.

This survey does not seek to unify the terminology, but we hope it will at least serve as a source to understand which terms mean the same and which terms are different.

\paragraph{Synergy} Methods from different communication approaches can benefit each other. For example, both the \type{sec:adversarial-examples}{adverserial examples} method \method{sec:adversarial-examples:hotflip}{HotFlip} and the \type{sec:counterfactuals}{counterfactual} method \method{sec:counterfactuals:mice}{MiCE} uses the \method{sec:input-features:gradient}{gradient w.r.t. the input} method from the \type{sec:input-features}{input feature} explanation literature. Recognizing these connections allows for flexibility in explanation methods. In the aforementioned example, other \type{sec:input-features}{input feature} explanations could have been used as well. Additionally, criticisms on the faithfulness of \type{sec:input-features}{input feature} methods could affect its dependents.

\paragraph{Helpful complex models} Models like GPT and T5 are immensely complex and thereby contribute to the interpretability challenge. However, importantly these models are not exclusively bad from an interpretability perspective, as they are also used to provide fluent explanations. For example, in \type{sec:counterfactuals}{counterfactual} explanations \method{sec:counterfactuals:polyjuice}{Polyjuice} uses the GPT-2 model and  \method{sec:counterfactuals:mice}{MiCE} uses the T5 model. Similarly, in \type{sec:natural-language}{natural language} explanations \method{sec:natural-language:cage}{CAGE} uses GPT. As such, these complex models can not be said to be exclusively counterproductive to interpretability.

%% file: chapters/future_directions.tex
\section{Future directions and challenges}
\label{sec:future-directions}

Interpretability for NLP is a fast-growing research field, with many methods being proposed each year. This survey provides an overview and categorization of many of these methods. In particular, we present \Cref{tab:overview} as a way to frame existing research. It is also the hope that \Cref{tab:overview} will help frame future research. In this section, we provide our opinions on what the most relevant challenges and future directions are in interpretability.

\paragraph{Measuring Interpretability} How interpretability is measured varies significantly. Throughout this paper, we have briefly documented how each method measures interpretability. A general observation is that each method paper often introduces its own measures of \measure{functionally-groundedness} or \measure{human-groundedness}. Even when established standards exist, such as the \emph{word intrusion test} \citep{Chang2009}, they get modified. This trend reduces comparability and risks invalidating the measure itself. 

It is important to recognize that measuring interpretability is, in some cases, inherently difficult. For example, in the case of measuring the \measure{functionally-groundedness} of \type{sec:input-features}{input feature} explanations, it is inherently impossible to provide gold labels for what is a correct explanation, because if humans could provide gold labels we wouldn’t need the explanation in the first place. This fundamentally leaves only proxy measures and axioms of \measure{functionally-groundedness}. However, this doesn’t mean highly principled proxy measures can’t be developed \citep{Hooker2019}.

For this reason, we are encouraging researchers and reviewers to value principled papers on measuring interpretability. Even if those measures don’t become established standards, a dedicated focus on measuring interpretability is a necessity for the integrity of the interpretability field.

\paragraph{Class explanations} There is a large number of papers on explanation methods. However, \category{class explanations} remain an underrepresented middle ground between \category{local} and \category{global explanations}.

The specific communication approach chosen should reflect its application, and for this reason, no explanation type can be said to be superior. However, it's important to recognize that \category{local explanations} can only provide anecdotal evidence and \category{global explanations} can be too abstract to ground what is explained. As such \category{class explanations} have their value, as they are not specific enough to be anecdotal. Simultaneously, they are grounded in the class they explain, making them easier to reason about. For this reason, we would encourage that \category{class explanation} gain equal representation in interpretability research.

\paragraph{Sequence-to-sequence explanations} In this survey we frequently comment on if a explanation method can be applied to sequence-to-class models or sequence-to-sequence models. Most methods are primarily made for sequence-to-class models, and the few that apply to sequence-to-sequence models are often not directly made for that purpose.

We suspect a reason for primarily explaining sequence-to-class models, is that sequence-to-sequence explanations may depend on interactive visualization to a greater extent \citep{Tenney2020, Strobelt2019, Madsen2019a}, which is harder to implement and write about in typical machine learning venues.

Regardless, sequence-to-sequence models are widely used in real-life applications, for example in machine translation. We therefore advocate for developing more explanations for sequence-to-sequence models, or at the very least include an evaluation on a sequence-to-sequence model in papers that provide methods that can operate on both types of models.

\paragraph{Combining post-hoc with intrinsic methods} \posthoc{Post-hoc} and \intrinsic{intrinsic} methods are in literature, including this paper, represented as distinct. However, there are important middle grounds. 

As mentioned in the introduction, most \intrinsic{intrinsic} methods are not purely intrinsic. They often have an intermediate representation, which can be intrinsically interpretable. However, producing this representation is often done with a black-box model. For this reason, \posthoc{post-hoc} explanations are needed if the entire model is to be understood.

Beyond this direction, there are works where the training objective and procedure helps to provide better \posthoc{post-hoc} explanations. This survey briefly argues that the \method{sec:input-features:shap}{Kernel SHAP} method exists in this middle ground, as it depends on input-masking being part of the training procedure. In computer vision, \citet{Bansal2020} show that adding noise to the input images creates better \type{sec:input-features}{input feature} explanations. In general, we hope to see more work in this direction.

%% file: chapters/conclusion.tex
\section{Conclusion}
This survey presents an overview of \posthoc{post-hoc} interpretability methods for neural networks in NLP. The main content of this survey is on the interpretability methods themself and how they communicate their explanation of the model. This content is categorized through \Cref{tab:overview}.

Throughout the survey, we also refer back to \measure{measures of interpretability} (\cref{sec:measures-of-interpretability}) to describe how each paper evaluates its proposed method. Measuring interpretability is an often undervalued aspect of interpretability with little standardization of the benchmarks. However, by briefly mentioning each method of measurement, we hope that this will lead to less fragmentation.

Finally, we discuss interesting findings and future directions, which we consider particularly important. Overall, we hope that \Cref{tab:overview}, the discussions of each communication approach and their methods, and the final discussion sections help frame future research and provide broad insight to those who apply interpretability.